\documentclass{article}

\usepackage{amsmath,amsfonts,bm}
\usepackage{iclr2027_conference,times}
\setcitestyle{numbers,square}
\usepackage{xcolor}
\usepackage{amsmath}
\usepackage{graphicx} 
\usepackage{enumitem}
\usepackage[table]{xcolor}
\usepackage{booktabs}
\usepackage{xspace} 
\usepackage{hyperref}
\usepackage{wrapfig}
\usepackage{tabularx}

\newcommand{\methodName}{\textsc{EMG-FM-Bench}\xspace}

\title{\methodName: A Comprehensive Benchmark for Foundation Model Transfer and Adaptation on Electromyography\\[1pt]}

\author{%
{\normalfont
\centering
\shortstack[c]{%
\textbf{Tianhao Wu}$^{1}$ \quad
\textbf{Jiawei Yu}$^{1}$ \quad
\textbf{Xiangdong Hu}$^{1}$ \quad
\textbf{Xu Wu}$^{2}$ \\[4pt]
\textbf{Hongyu Wu}$^{2}$ \quad
\textbf{Sidharth Anupkrishnan}$^{3}$ \quad
\textbf{Amirmohammad Radmehr}$^{3}$ \\[4pt]
\textbf{Yi Chen Liu}$^{1}$ \quad
\textbf{Phuc Nguyen}$^{3}$ \quad
\textbf{Yi Wu}$^{2}$ \quad
\textbf{Jian Liu}$^{1}$ \\[9pt]
{\small
$^{1}$University of Georgia \quad
$^{2}$University of Oklahoma \quad
$^{3}$University of Massachusetts Amherst
} \\[7pt]
{\footnotesize
\texttt{\{Tianhao.Wu1,Jiawei.Yu,Easton.Hu,YiChen.Liu,jianliu\}@uga.edu}
} \\[3pt]
{\footnotesize
\texttt{\{Xu.Wu-1,lidowu,Yi.Wu-1\}@ou.edu}
\quad
\texttt{\{sanupkrishna,aradmehr,phuc\}@umass.edu}
}
}%
}%
}

\iclrfinalcopy

\begin{document}

\maketitle

\begin{abstract}

Foundation models (FMs) are increasingly being developed for general time series and physiological signals, yet their transferability to downstream physiological tasks remains poorly understood. This question is particularly challenging for electromyography (EMG), where signal distributions vary substantially across users, sensing configurations, acquisition hardware, and downstream tasks. We introduce \methodName, a systematic benchmark for studying foundation-model transfer and adaptation on EMG. \methodName{} unifies 20 public datasets with over 1 million EMG segments and evaluates nine pretrained foundation models across four questions: how pretrained models perform when frozen or fully fine-tuned, how much pretraining helps compared with training the same model from scratch, how well models generalize to new users with limited labeled data, and how performance changes across different EMG tasks. Across the benchmark, linear probing provides useful information about pretrained representations, but full fine-tuning can substantially change downstream EMG performance. Comparing each pretrained model with the same model trained from scratch shows that the benefit of pretraining varies substantially across models and is not universal. Performance decreases when models are evaluated on new users, while five-shot adaptation improves macro-F1 in 70.2\% of evaluated model--dataset combinations but recovers only part of the lost performance. Model performance is highly consistent between upper- and lower-limb classification and remains strongly correlated with continuous EMG-to-text decoding. Together, these results provide a systematic view of when pretrained time-series models transfer effectively to EMG and how their performance depends on fine-tuning, user variation, and downstream task.

\end{abstract}

\section{Introduction}
\label{sec:introduction}


Time-series foundation models are increasingly positioned as reusable backbones for downstream temporal tasks. Their transfer behavior beyond the pretraining objective, however, remains poorly understood. Recent studies suggest that downstream utility can differ substantially from what is implied by pretraining performance alone, and that larger model scale does not necessarily translate into better transfer~\cite{zhang2026unified,yang2026eeg}. EMG provides a particularly stringent setting in which to examine these questions. EMG distributions vary with subject physiology, electrode placement, acquisition hardware, and recording session~\cite{huang2026repurposing,shi2024emg,ghislieri2021long,yang2024emgbench}. Public EMG datasets additionally span widely different sampling rates, channel configurations, and task formulations, from discrete motor-intent recognition and locomotion analysis to continuous decoding~\cite{al2023electromyography,sivakumar2024emg2qwerty,gonzalez2020silent}. It therefore remains unclear whether pretrained time-series models can transfer effectively to EMG across variations in users, sensing setups, and downstream tasks, and how much downstream adaptation is required to make them useful.


Existing EMG evaluation efforts have progressively expanded the study of robustness across trials, users, and sessions. Prior work has examined trial-invariant gesture recognition~\cite{dai2023improved}, contributed standardized datasets and processing resources~\cite{jiang2021open}, compared cross-user adaptation strategies~\cite{hoshino2024comparison}, and evaluated cross-subject generalization and limited user calibration across multiple gesture datasets~\cite{yang2024emgbench}. These studies primarily focus on whether task-specific EMG models can generalize or adapt under changes in users and recording conditions. Pretrained time-series foundation models introduce a broader transfer problem: models developed with different architectures, pretraining objectives, and source domains must be adapted to EMG, where sensing characteristics and downstream requirements can differ substantially from those seen during pretraining. However, existing EMG benchmarks do not systematically evaluate the transfer and adaptation behavior of general time-series foundation models on EMG across different evaluation settings, users, and downstream tasks.

\begin{figure*}[t]
\vspace{-6mm}
\centering
\includegraphics[width=0.75\textwidth]{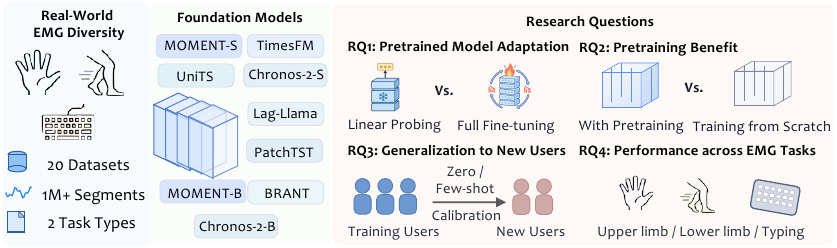}
\caption{
    \methodName{} evaluates nine pretrained models on 20 heterogeneous EMG datasets under four research questions (RQ1--RQ4).
}
\label{fig:overview}
\vspace{-4mm}
\end{figure*}

To address this gap, we introduce~\methodName, a systematic benchmark for studying foundation-model transfer and adaptation on EMG. As shown in Figure~\ref{fig:overview}, \methodName{} evaluates nine pretrained foundation models across 20 public datasets with over 1 million EMG segments under a unified evaluation framework, spanning diverse architectures, pretraining objectives, model scales, and source domains, with representative examples including MOMENT~\cite{goswami2024moment}, Chronos-2~\cite{ansari2025chronos}, and BRANT~\cite{zhang2023brant}.
We organize the benchmark around four complementary research questions that examine foundation-model transfer and adaptation from different perspectives: 

\textbf{RQ1: Pretrained Model Adaptation.}
\textit{How useful are pretrained models for EMG when the backbone is kept frozen, and how does full fine-tuning change their performance?} A central goal of foundation-model pretraining is to learn representations that can be reused for downstream tasks; we therefore compare linear probing, which keeps the pretrained backbone frozen, with full fine-tuning of the entire model. 

\textbf{RQ2: Pretraining Benefit.}
\textit{How much benefit does pretraining provide for downstream EMG performance beyond the underlying architecture?} Strong performance after full fine-tuning does not necessarily imply that pretraining itself is beneficial, so we compare pretrained models with architecture-matched models trained from scratch. 

\textbf{RQ3: Generalization to New Users.}
\textit{How well do pretrained models generalize to users unseen during downstream training, and how much can limited labeled data improve their performance?} Because EMG varies substantially across individuals, we evaluate held-out-user transfer, leave-one-subject-out (LOSO) generalization, and five-shot adaptation. 

\textbf{RQ4: Performance Consistency across EMG Tasks.} \textit{Is model performance consistent across different EMG tasks?} Because foundation models are intended to support diverse downstream tasks, we evaluate their behavior across upper-limb classification, lower-limb classification, and continuous EMG-to-text decoding.



Our main findings and contributions are:

\begin{enumerate}[leftmargin=*, label=\textbf{(\arabic*)}]
\vspace{-2mm}

\item \textbf{Full fine-tuning substantially changes pretrained-model performance (RQ1).}
Linear probing provides useful information about frozen representations, but does not fully reflect how the same models perform after end-to-end fine-tuning.

\item \textbf{Pretraining does not provide a universal advantage (RQ2).}
Architecture-matched comparisons show that pretrained initialization improves some models and datasets, while others perform similarly or better when trained from scratch.

\item \textbf{Cross-user generalization remains challenging (RQ3).}
Performance declines for unseen users, while limited labeled data often recovers part of the loss but does not remove substantial user-to-user variability.

\item \textbf{Model performance is consistent across EMG tasks (RQ4). }
Performance is strongly correlated across upper-limb classification, lower-limb classification, and continuous EMG-to-text decoding, although the strongest agreement occurs between the two classification categories.

\item \textbf{Open and reproducible evaluation.}
We release the standardized datasets, preprocessing pipelines, and evaluation code at
\url{https://anonymous.4open.science/r/EMG-FoundBench/} and
\url{https://huggingface.co/datasets/AnonymousAuthorSubmission/EMG-FoundBench}.
    
\end{enumerate}

\section{Background and Related Work}
\vspace{-2mm}

\subsection{FMs for Time-Series and Physiological Signals}
\vspace{-2mm}

Foundation-model pretraining has increasingly been extended from language~\cite{brown2020language,touvron2023llama,chowdhery2023palm} to time-series and physiological signals.

\noindent\textbf{General Time-Series FMs.} Recent models such as MOMENT~\cite{goswami2024moment}, Chronos~\cite{ansari2024chronos}, TimesFM~\cite{das2023decoder}, Moirai~\cite{woo2024unified}, Lag-Llama~\cite{rasul2310lag}, TimesNet~\cite{wu2022timesnet}, and ForecastPFN~\cite{dooley2023forecastpfn} learn transferable temporal representations from heterogeneous time-series corpora. However, these models were developed primarily for general-purpose forecasting, and their transferability to high-frequency, subject-dependent EMG remains unclear.

\noindent\textbf{Physiological Signal FMs.} Domain-specific foundation models have also emerged for ECG~\cite{choi2023ecgbert}, EEG/iEEG~\cite{jiang2024large,zhang2023brant}, and PPG~\cite{chen2025gpt,pillai2024papagei}. Recent EMG-specific models further explore large-scale neuromuscular representation learning~\cite{kurbis2025emg,fasulo2025tinymyo,huang2026reading}. Nevertheless, EMG evaluation remains fragmented across datasets, preprocessing pipelines, model availability, and adaptation settings. Its sensitivity to non-stationarity~\cite{phinyomark2018emg,reaz2006techniques}, anatomy~\cite{dietz1979neuronal}, electrode placement, and subject variation motivates a standardized evaluation of when pretrained temporal representations transfer to EMG and how much downstream adaptation they require. Complementary to EMG-specific models, we evaluate broadly pretrained time-series models across heterogeneous EMG datasets and tasks.
\subsection{Time-Series and EMG Evaluation Benchmarks}
\vspace{-2mm}

TSFM-Bench~\cite{li2025tsfm} and GIFT-Eval~\cite{aksu2024gift} evaluate time-series FMs under zero-, few-, and full-shot settings, but primarily on general-purpose forecasting datasets. Brain4FMs~\cite{shen2026brain4fms} extends FM benchmarking to EEG and iEEG, while recent PPG benchmarks compare generalist and specialist models for cardiovascular applications~\cite{kataria2025generalist}. These benchmarks do not address the distinct cross-user and task-format challenges of peripheral neuromuscular signals.

Early EMG benchmarks were primarily dataset-centric. NinaPro introduced a family of benchmark databases for upper-limb gesture recognition~\cite{atzori2014electromyography}, while Hyser provided a high-density EMG dataset, an analysis toolbox, and reference results for gesture recognition and force estimation~\cite{jiang2021open}. EMGBench~\cite{yang2024emgbench} subsequently established the first open-source multi-dataset benchmark for EMG generalization and adaptation, standardizing nine upper-limb gesture datasets under leave-one-subject-out, few-shot, and intersession protocols. Its central question is how an EMG gesture classifier trained on source users generalizes and adapts to a new user. To this end, it compares EMG-to-image representations and ImageNet-pretrained CNN and Transformer backbones, with all model parameters updated during training. The pretrained backbone is therefore treated as a component of the downstream classifier rather than as the object of evaluation: frozen representation reuse is not examined, and no matched pretrained-versus-random-initialization results are reported.


\subsection{Our Positioning}
\vspace{-2mm}

\methodName{} shifts the evaluation question from whether a task-specific EMG classifier can adapt across users to what transfers when general-purpose pretrained time-series models are repurposed for EMG. To our knowledge, it is the first benchmark to evaluate this question through matched representation- and initialization-level controls. Linear probing measures whether EMG-relevant information is directly accessible from frozen features, full fine-tuning measures adaptation capacity, and architecture-matched random initialization provides an architecture-matched estimate of the benefit associated with pretrained initialization beyond the model design. Across 20 datasets and nine model configurations, we further examine whether these conclusions persist under subject shift and across upper-limb recognition, lower-limb locomotion, and continuous EMG-to-text decoding. The benchmark is therefore not another evaluation of cross-user adaptation alone, but a diagnostic framework for determining when pretrained temporal representations transfer to EMG, where their gains originate, and whether those gains generalize across evaluation regimes and tasks.

\section{Evaluation Methodology}
\vspace{-3mm}
\subsection{Datasets}
\vspace{-2mm}
FMs must generalize across diverse data distributions, task formats, and sensor configurations. Homogeneous benchmarks cannot test these capabilities. We curate 20 public EMG datasets spanning different body regions, hardware setups, task types, and modalities (Appendix~\ref{app:datasets}). The benchmark includes 15 upper-limb gesture-recognition datasets and four lower-limb locomotion datasets. Classification tasks (19 datasets) and sequence-to-sequence decoding (1 dataset) probe task flexibility. Sampling rates from 100 to 2400~Hz and channel counts from 6 to 128 challenge variation across channel configurations. Our benchmark has three categories:

\noindent\textbf{Upper-Limb Gesture Datasets (15 datasets):}
The upper-limb group includes the NinaPro family~\cite{atzori2014electromyography,pizzolato2017comparison,palermo2017repeatability,krasoulis2017improved}, GRABMyo~\cite{pradhan2022multi}, CapgMyo~\cite{dai2021capg}, Pinch and Roshambo~\cite{donati2018processing}, Arm Translation~\cite{kyranou2025emg}, emg2pose~\cite{salter2024emg2pose}, and 3DC and ContinuousTrans~\cite{eddy2023libemg}. These datasets capture hand and forearm muscle activations during controlled gesture tasks with varying sampling rates (100--2048 Hz) and channel densities (6--128 channels), covering 3--52 gesture classes across datasets.

\begin{wraptable}[10]{r}{0.60\textwidth}
\vspace{-2mm}
\centering
\small

\caption{Scale and task coverage of \methodName.
Dataset-level statistics are provided in
Appendix~\ref{app:datasets}.}
\label{tab:datasets_summary}

\setlength{\tabcolsep}{3.2pt}
\renewcommand{\arraystretch}{1.12}

\resizebox{\linewidth}{!}{
\begin{tabular}{lrrrl}
\toprule
\textbf{Category} &
\textbf{\# Data} &
\textbf{\# Subj.} &
\textbf{\# Windows} &
\textbf{Benchmark Task} \\
\midrule
Upper limb & 15 & 489 & 574,182 & Gesture/pose classification \\
Lower limb & 4 & 156 & 54,953 & Locomotion classification \\
Typing & 1 & 108 & 402,119 & Character sequence decoding \\
\midrule
\textbf{Total} &
\textbf{20} &
\textbf{753} &
\textbf{1,031,254} &
Classification and Seq2Seq \\
\bottomrule
\end{tabular}
}
\end{wraptable}

\noindent\textbf{Lower-Limb Locomotion Datasets (4 datasets):}
HD-EMG~\cite{dimitrov2023high}, SIAT-LLMD~\cite{wei2023surface}, Mendeley Knee~\cite{mohr2019classification}, and MyPredict~\cite{schulte2023roessingh} capture muscle activity during gait and locomotion tasks, extending the benchmark beyond upper-limb recognition. HD-EMG additionally provides synchronized IMU, kinematic, and kinetic measurements; these auxiliary modalities are treated as dataset attributes rather than as a separate benchmark category, and the present evaluation uses EMG inputs.

\noindent\textbf{Sequence-to-Sequence Dataset (1 dataset):} emg2qwerty~\cite{sivakumar2024emg2qwerty}, a large-scale typing dataset with bilateral wrist EMG (32 channels, 2000 Hz) from 108 subjects performing natural keyboard typing across 5,262,671 keypress events. This enables continuous decoding evaluation using CTC loss and character error rate.

Table~\ref{tab:datasets_summary} summarizes key statistics. Complete dataset descriptions, preprocessing protocols, and task specifications are provided in Appendix~\ref{app:datasets}.


\vspace{-3mm}
\subsection{Data Processing}
\vspace{-2mm}
EMG signals contain physiological information in the 20-450 Hz range but suffer from motion artifacts (lower than 20 Hz), electrical noise (higher than 450 Hz), and powerline interference (50/60 Hz)~\cite{merletti2016surface}. To provide a consistent temporal interface across foundation models, we map all recordings to a common 1000-Hz grid using polyphase resampling. This operation incorporates low-pass filtering during rate
conversion and ensures that windows of the same physical duration have comparable sequence lengths and computational costs across datasets. For recordings acquired below 1000 Hz, resampling only harmonizes the temporal grid and does not recover frequency components above the Nyquist frequency
of the original acquisition. For recordings sampled below 1000 Hz, we apply bandpass filtering at the native sampling rate, with the upper cutoff constrained by the original Nyquist frequency, before upsampling; recordings sampled at or above 1000 Hz are resampled using polyphase anti-aliasing and then filtered at 20--450 Hz. A 50/60-Hz notch filter is applied where the corresponding frequency is supported by the original sampling rate, followed by per-channel, per-window (instance-level) Z-score normalization~\cite{besomi2019consensus}. To provide a common temporal interface across models, we use non-overlapping 3-second windows for all downstream tasks (with the exception of CapgMyo and HD-EMG), which also standardizes the temporal context and computational budget across datasets and models. For classification, we first partition recordings at the trial or repetition level and then divide each partition into consecutive non-overlapping windows, assigning each window its corresponding discrete label. For emg2qwerty, we similarly segment each recording into consecutive non-overlapping 3-second windows; the timestamped keypresses occurring within each window form a variable-length target sequence, and the models are trained using CTC loss~\cite{graves2006connectionist}. Appendix~\ref{app:datasets} provides further preprocessing and segmentation details.



\vspace{-4mm}
\subsection{Foundation Model Baselines}
\vspace{-2mm}

\begin{table*}[t]
\centering
\caption{Pretrained model families evaluated in \methodName{}. Models span different Transformer architectures, pretraining domains, and parameter scales. Parameter counts correspond to the evaluated checkpoints.}
\label{tab:model_summary1}
\small
\setlength{\tabcolsep}{5pt}
\renewcommand{\arraystretch}{1.05}

\begin{tabularx}{\textwidth}{
>{\raggedright\arraybackslash}p{3.0cm}
>{\raggedright\arraybackslash}p{4.0cm}
>{\raggedright\arraybackslash}X
>{\centering\arraybackslash}p{2.2cm}
}
\toprule
\textbf{Model} &
\textbf{Architecture} &
\textbf{Pretraining Domain} &
\textbf{Params.} \\
\midrule

MOMENT-S/B~\cite{goswami2024moment}
& Encoder-only Transformer
& Heterogeneous time series
& 40M / 125M \\

Chronos-2-S/B~\cite{ansari2025chronos}
& Encoder-only Transformer
& General forecasting data
& 28M / 120M \\

TimesFM~\cite{das2023decoder}
& Decoder-only Transformer
& General time series
& 200M \\

Lag-Llama~\cite{rasul2310lag}
& Decoder-only Transformer
& General time series
& 2.45M \\

PatchTST~\cite{nie2022time}
& Encoder-only Transformer
& General time series
& $\sim$1M \\

UniTS~\cite{gao2024units}
& Modified Transformer
& Heterogeneous time series
& 3.4M \\

BRANT~\cite{zhang2023brant}
& Temporal--spatial Transformer
& Intracranial EEG
& $>$500M \\

\bottomrule
\end{tabularx}
\end{table*}

Table~\ref{tab:model_summary1} summarizes the seven model families and nine pretrained configurations evaluated in \methodName{}. The selection spans diverse Transformer architectures, parameter scales from approximately 1M to over 500M, and both general-purpose and electrophysiology-specific pretraining. The models also differ in their original pretraining objectives, including masked reconstruction, forecasting, and multi-task learning. The MOMENT and Chronos-2 variants additionally enable within-family comparisons across model scales. For EMG classification, we attach a linear classification head to the representation produced by each backbone and evaluate all models under the same downstream protocols. Model-specific input adaptations, architectural details, and hyperparameters are provided in Appendix~\ref{app:models}.

\noindent\textbf{General-Purpose Time-Series Models:}
We evaluate eight configurations from six general-purpose time-series model families. MOMENT~\cite{goswami2024moment} and PatchTST~\cite{nie2022time} provide encoder-only Transformer backbones, while Chronos-2~\cite{ansari2025chronos} uses an encoder-only architecture with temporal and group attention. TimesFM~\cite{das2023decoder} and Lag-Llama~\cite{rasul2310lag} represent decoder-only forecasting models, whereas UniTS~\cite{gao2024units} uses a modified Transformer architecture designed for multiple time-series tasks. We evaluate two MOMENT variants, Small (40M parameters) and Base (125M), and two Chronos-2 variants, Small (28M) and Base (120M); the remaining models range from  $\sim$1M parameters for PatchTST to 200M for TimesFM. Their pretraining corpora span heterogeneous time-series domains, including physiological signals in some cases, and differ substantially across model families.

\noindent\textbf{Electrophysiology-Specific Model:}
BRANT~\cite{zhang2023brant} has over 500M parameters and uses sequential temporal and spatial Transformer encoders. It was pretrained on approximately 1.01 TB of intracranial EEG recordings and provides an electrophysiology-specific counterpoint to the general-purpose time-series models. Its released weights enable us to evaluate this pretraining background under the same EMG transfer protocol.

Together, these models span architecture families, parameter scales, and pretraining domains under a common downstream evaluation framework. Appendix~\ref{app:models} provides implementation details.




\vspace{-1mm}
\section{Results}
\label{sec:results}
\vspace{-2mm}

\subsection{Evaluation Protocols and Benchmark Overview}
\label{sec:protocols}

We evaluate the models under six protocols corresponding to the four research questions introduced in Section~\ref{sec:introduction}. For \textbf{RQ1}, linear probing (LP) freezes the pretrained backbone and trains only the classification head, while full fine-tuning (FFT) updates the entire model. Their comparison measures how model performance changes when the pretrained backbone is fully adapted to EMG. For \textbf{RQ2}, each pretrained model is compared with the same model trained from random initialization under the same downstream setup, allowing us to measure the benefit associated with pretraining.

For \textbf{RQ3}, models trained on Pool A are first evaluated directly on the held-out Pool B users without calibration. We additionally use leave-one-subject-out (LOSO) evaluation to measure variation across individual held-out users. Five-shot adaptation then updates the Pool A model with five labeled examples per class drawn from the training split of the held-out Pool B users and evaluates on the Pool B test split. Because these adaptation examples are pooled across the held-out users rather than collected separately from each one, this protocol asks whether a small amount of labeled data from the new-user population is enough to recover the performance lost under user shift, rather than how much adaptation an individual user needs. For \textbf{RQ4}, we compare model behavior across upper-limb classification, lower-limb classification, and continuous EMG-to-text decoding.
\vspace{-4mm}
\paragraph{Protocol controls.}
For a given model, the input interface, representation aggregation, classification head, data splits, and checkpoint-selection procedure are held fixed across the corresponding protocol comparisons. LP freezes the backbone, whereas FFT updates it. The scratch setting uses the same model architecture and downstream interface as FFT but initializes the backbone randomly. Models are trained for up to 50 epochs with validation-based early stopping, except for five-shot adaptation, which uses the fixed adaptation procedure described in Appendix~\ref{app:training}. Complete implementation details for the model-specific downstream interfaces are provided in Appendix~\ref{app:classification_interfaces}.

\paragraph{Aggregation and metrics.}
Benchmark-level analyses use all 19 classification datasets. We report accuracy and macro-F1 for performance comparisons and use paired differences between protocols when evaluating adaptation, pretraining benefit, and new-user generalization. Mean ranks are used only as a complementary measure when comparing relative model behavior across datasets with different class counts and difficulty levels. Complete per-dataset results are reported in Appendix~\ref{app:complete_results}.

\begin{table*}[t]
\centering
\caption{Aggregate results across 19 classification datasets. 
Each cell reports the mean score with mean rank in parentheses; lower rank is better. The 5-shot column reports macro-F1 averaged over five seeds; all other columns report accuracy.}
\label{tab:master}

\footnotesize
\setlength{\tabcolsep}{3.5pt}
\renewcommand{\arraystretch}{0.95}

\begin{tabular}{lcccccc}
\toprule
\textbf{Model} & \textbf{LP} & \textbf{FFT} & \textbf{Scratch} &
\textbf{Held-out} & \textbf{LOSO} & \textbf{5-shot (F1)} \\
\midrule
MOMENT-S    & 41.9 (2.5) & 60.1 (2.1) & 57.4 (2.1) & 39.0 (3.3) & 34.9 (3.8) & 42.7 (2.6) \\
MOMENT-B    & 52.2 (1.2) & 46.8 (4.4) & 57.3 (2.1) & 28.1 (5.7) & 35.6 (3.9) & 40.0 (3.5) \\
BRANT       & 18.2 (7.5) & 22.8 (8.0) & 20.8 (8.5) & 19.8 (7.6) & 23.0 (6.9) & 12.5 (8.3) \\
PatchTST    & 20.2 (7.1) & 30.1 (7.4) & 30.7 (6.9) & 27.2 (5.6) & 19.6 (8.6) & 16.3 (7.1) \\
TimesFM     & 29.4 (5.2) & 37.9 (5.9) & 31.6 (6.8) & 31.1 (5.0) & 34.2 (4.6) & 27.6 (5.3) \\
Chronos-2-S & 28.1 (4.6) & 54.3 (3.1) & 38.2 (4.8) & 41.6 (3.2) & 36.7 (3.6) & 44.0 (2.9) \\
Chronos-2-B & 29.4 (4.1) & 56.5 (2.8) & 40.4 (4.1) & 43.8 (2.6) & 38.5 (2.8) & 45.5 (2.6) \\
Lag-Llama   & 21.6 (7.0) & 38.0 (6.3) & 42.6 (4.3) & 28.4 (6.0) & 31.0 (5.6) & 16.7 (7.5) \\
UniTS       & 26.2 (5.8) & 40.2 (5.0) & 37.9 (5.4) & 27.7 (6.0) & 31.2 (5.1) & 27.2 (5.2) \\
\bottomrule
\end{tabular}

\vspace{-2mm}
\end{table*}

Table~\ref{tab:master} provides an overview of how model performance changes across the six evaluation protocols. The differences between LP and FFT motivate the analysis of pretrained-model adaptation in RQ1, while the contrast between FFT and scratch is examined in RQ2. The lower held-out-user performance and changes after five-shot adaptation motivate the new-user analysis in RQ3. RQ4 then examines whether the relative behavior observed on classification tasks remains consistent across different EMG task types. Mendeley Knee is a participant-level classification task (injury history); its held-out and LOSO results are therefore the most informative for this dataset (Appendix~\ref{sec:mendeley}).

\paragraph{Conventional supervised reference.}
We additionally include a randomly initialized BiLSTM as a representative generic supervised temporal reference under the scratch, held-out-user, LOSO, and five-shot protocols. Architectural details are provided in Appendix~\ref{app:lstm}, with complete results reported in Appendices~\ref{app:pretraining_contribution}
and~\ref{app:cross_user}.
\vspace{-2mm}
\subsection{RQ1: Pretrained Model Adaptation}
\label{sec:rq1}

RQ1 examines how much EMG performance changes when a pretrained backbone is fully fine-tuned rather than kept frozen. Complete full fine-tuning and linear-probing results across all 19 classification datasets are reported in Appendix Tables~\ref{tab:app_downstream_all} and~\ref{tab:app_full_finetune_all}.

Across the 171 model--dataset pairs, full fine-tuning improves accuracy in 136 cases (79.5\%) and macro-F1 in 141 cases (82.5\%), with average gains of 13.3 and 16.4 percentage points, respectively. However, the effect varies substantially across backbones. As shown in Table~\ref{tab:rq1_adaptation_gain}, Chronos-2-S and Chronos-2-B improve macro-F1 on all 19 datasets, whereas MOMENT-B improves macro-F1 on only six datasets and decreases by 5.4 accuracy points on average. UniTS gains 14.0 accuracy points on average and improves macro-F1 on 18 of 19 datasets, showing that it benefits substantially from end-to-end adaptation. Thus, full fine-tuning is beneficial in most model--dataset combinations, but neither the magnitude nor the direction of the change is uniform across pretrained models.

\begin{wraptable}{r}{0.47\columnwidth}
\centering
\caption{Effect of full fine-tuning relative to linear probing across the 19 classification datasets. 
$\Delta$Acc denotes $\mathrm{Acc}_{\mathrm{FFT}}-\mathrm{Acc}_{\mathrm{LP}}$.}
\label{tab:rq1_adaptation_gain}
\scriptsize
\setlength{\tabcolsep}{3.5pt}
\renewcommand{\arraystretch}{1.05}

\begin{tabular}{lrr}
\toprule
\textbf{Model} &
\textbf{Mean $\Delta$Acc} &
\textbf{Improved (F1)} \\
\midrule
MOMENT-S    & +18.2 & 18/19 \\
MOMENT-B    &  -5.4 &  6/19 \\
BRANT       &  +4.6 & 14/19 \\
PatchTST    &  +9.9 & 18/19 \\
TimesFM     &  +8.5 & 15/19 \\
Chronos-2-S & +26.2 & 19/19 \\
Chronos-2-B & +27.1 & 19/19 \\
Lag-Llama   & +16.4 & 14/19 \\
UniTS       & +14.0 & 18/19 \\
\bottomrule
\end{tabular}

\vspace{-2mm}
\end{wraptable}

We next ask whether performance under linear probing is nevertheless informative about model behavior after full fine-tuning. At the aggregate level, LP and FFT mean ranks remain strongly associated, with Spearman correlations of $\rho=0.88$ for accuracy and $\rho=0.82$ for macro-F1, suggesting that frozen evaluation preserves broad differences between stronger and weaker configurations. However, this agreement is much weaker at the individual-dataset level: the model with the highest LP accuracy is also the highest-performing FFT model on only 1 of 19 datasets, and the corresponding agreement is 2 of 19 datasets by macro-F1. Thus, frozen evaluation can provide a coarse indication of overall model strength, but it does not reliably identify which model will perform best after full adaptation on a specific EMG dataset.

Overall, full fine-tuning improves EMG performance for most model--dataset pairs, but the magnitude and even direction of the change are highly model-dependent. Linear probing captures broad performance differences across models, yet it is not a reliable substitute for evaluating models after end-to-end adaptation.
\vspace{-1mm}
\vspace{-1mm}
\subsection{RQ2: Pretraining Benefit}
\label{sec:rq2}

\vspace{-1mm}
RQ2 asks how much pretraining improves downstream EMG performance beyond what can be achieved by training the same model architecture from scratch. For each model, we compare pretrained full fine-tuning with a randomly initialized counterpart using the same downstream architecture, interface, data split, and training protocol. We define
$\Delta\mathrm{Acc}=\mathrm{Acc}_{\mathrm{FFT}}-\mathrm{Acc}_{\mathrm{Scratch}}$.
Complete per-dataset FFT and scratch results are reported in Appendix Tables~\ref{tab:app_downstream_all} and~\ref{tab:app_scratch_all}. Our goal is not to estimate the best achievable performance of each model, but to compare how pretrained and randomly initialized versions of the same architecture behave under the same downstream adaptation procedure.

\begin{wrapfigure}{r}{0.6\columnwidth}
    \vspace{-3mm}
    \centering
    \includegraphics[width=\linewidth]{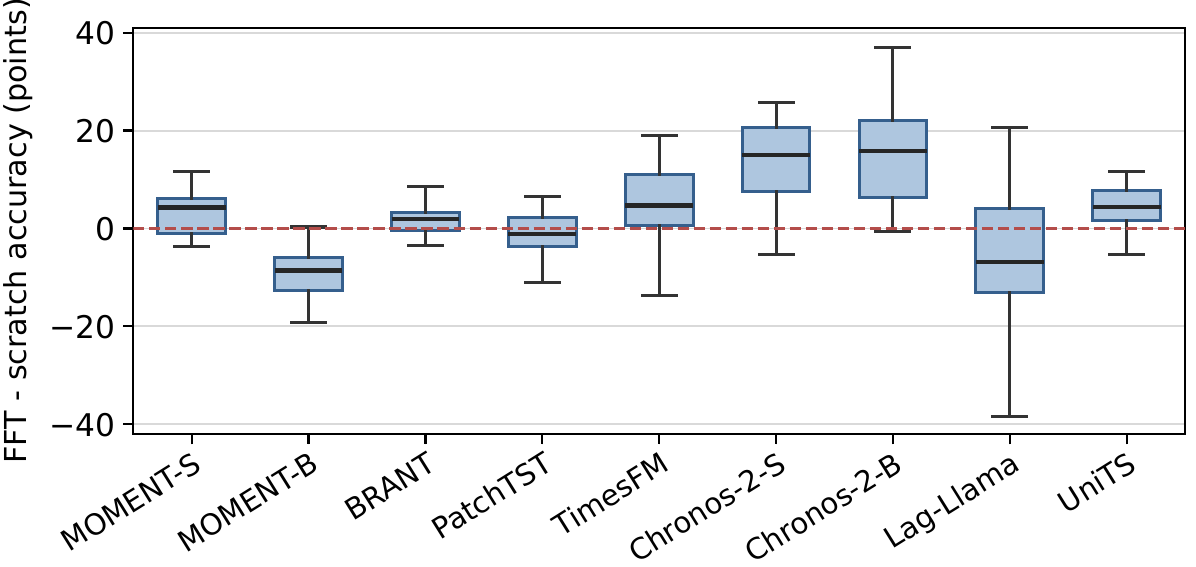}
    \caption{Accuracy difference between pretrained full fine-tuning and training the same model from scratch across 19 classification datasets. Positive values indicate a benefit from pretraining.}
    \label{fig:rq2_pretraining_gain}
    \vspace{-4mm}
\end{wrapfigure}

Across the 171 model--dataset comparisons, pretrained models outperform their scratch counterparts in 106 cases (62.0\%) by accuracy and 100 cases (58.5\%) by macro-F1. The overall mean differences are +3.3 percentage points in accuracy and +3.0 points in macro-F1. Pretraining therefore provides a modest average benefit across the benchmark, but the effect is far from universal and varies substantially across model families (Figure~\ref{fig:rq2_pretraining_gain}), with model-level mean accuracy differences ranging from $-10.5$ to $+16.1$ points.

This variation is particularly clear across individual model families. Chronos-2-S and Chronos-2-B benefit consistently from pretraining, outperforming their scratch counterparts in accuracy on 17 and 18 of 19 datasets and gaining 16.1 and 16.0 points on average, respectively. TimesFM also shows a positive average gain of 6.3 points. UniTS obtains a smaller average gain of 2.3 points and improves over scratch on 15 of 19 datasets by both accuracy and macro-F1. In contrast, MOMENT-B, PatchTST, and Lag-Llama obtain lower average accuracy after pretrained fine-tuning than when trained from scratch. These results show that the benefit of pretraining is strongly model-dependent rather than a general property of foundation-model initialization.

Importantly, strong downstream performance after fine-tuning does not necessarily imply a large benefit from pretraining. MOMENT-S achieves strong FFT performance but gains only 2.7 accuracy points over its scratch counterpart, whereas the two Chronos-2 variants gain approximately 16 points. Across the nine model configurations, pretraining gain is not strongly associated with absolute FFT performance (Spearman $\rho=0.43$). Thus, absolute downstream performance and benefit from pretraining capture different aspects of model behavior.

\paragraph{Effect of model scale.}
The two within-family comparisons also show that larger checkpoints do not consistently obtain larger pretraining benefits. MOMENT-S gains 2.7 points on average, whereas MOMENT-B decreases by 10.5 points. In contrast, Chronos-2-S and Chronos-2-B obtain nearly identical gains of 16.1 and 16.0 points. These four checkpoints are insufficient to establish a general scaling relationship, but they show that increased parameter count does not automatically translate into greater EMG transfer benefit.

\vspace{-1mm}

\vspace{-1mm}
\paragraph{Conventional supervised reference.}
The BiLSTM (approximately 2.4M parameters) remains competitive with the much larger Transformer backbones when all models are trained from scratch. It falls within three accuracy points of the best scratch configuration on Roshambo and MyPredict, achieves the highest macro-F1 among all scratch models on MyPredict, and outperforms the scratch variants of BRANT, PatchTST, and TimesFM on 16, 11, and 10 of the 19 datasets, respectively (Appendix Table~\ref{tab:app_scratch_all}). Architectural scale alone therefore does not translate into better EMG classification performance in the absence of useful pretraining.

In summary, pretraining provides substantial benefits for some model families, modest benefits for others, and lower performance than scratch under the shared adaptation protocol for several architectures. High downstream performance after fine-tuning should therefore not be interpreted as evidence that pretraining itself was beneficial, and larger model scale does not consistently improve EMG performance, as the competitive BiLSTM reference also demonstrates.
\vspace{-2mm}
\subsection{RQ3: Generalization to New Users}
\label{sec:rq3}

RQ3 examines how well pretrained models generalize to users not seen during downstream training, how much performance varies across individual users, and how much limited labeled data from new users can recover performance. Complete held-out-user, LOSO, and five-shot results are reported in Appendix Tables~\ref{tab:app_zero_shot_all}, \ref{tab:app_loso_all}, and~\ref{tab:app_few_shot_all}, respectively.

\begin{wraptable}{r}{0.54\columnwidth}
\vspace{-2mm}
\centering
\caption{Model-level new-user degradation and five-shot recovery across 19 classification datasets (macro-F1).
$\Delta_{\mathrm{New}}=\mathrm{F1}_{\mathrm{Held\text{-}out}}-\mathrm{F1}_{\mathrm{FFT}}$ and
$\Delta_{\mathrm{5shot}}=\mathrm{F1}_{\mathrm{5shot}}-\mathrm{F1}_{\mathrm{Held\text{-}out}}$, where $\mathrm{F1}_{\mathrm{5shot}}$ is averaged over five seeds.
\textit{Drop} and \textit{Gain} denote the number of datasets with decreased held-out macro-F1 and improved five-shot macro-F1, respectively.}
\label{tab:rq3_generalization}

\scriptsize
\setlength{\tabcolsep}{3pt}
\renewcommand{\arraystretch}{1.05}

\begin{tabular}{lrrrr}
\toprule
\textbf{Model} &
\textbf{$\Delta_{\mathrm{New}}$} &
\textbf{Drop} &
\textbf{$\Delta_{\mathrm{5shot}}$} &
\textbf{Gain} \\
\midrule
MOMENT-S    & -27.1 & 18/19 & +11.9 & 16/19 \\
MOMENT-B    & -22.4 & 18/19 & +18.3 & 19/19 \\
BRANT       &  -2.5 & 13/19 &  +1.2 & 13/19 \\
PatchTST    &  -6.3 & 15/19 &  -3.5 &  6/19 \\
TimesFM     & -10.4 & 15/19 &  +4.2 & 13/19 \\
Chronos-2-S & -19.1 & 18/19 & +10.9 & 15/19 \\
Chronos-2-B & -19.0 & 19/19 &  +9.9 & 16/19 \\
Lag-Llama   & -12.0 & 12/19 &  -2.5 &  9/19 \\
UniTS       & -15.3 & 18/19 &  +6.1 & 13/19 \\
\bottomrule
\end{tabular}

\vspace{-2mm}
\end{wraptable}

\vspace{-2mm}
\paragraph{Generalization to new users.}
We compare in-distribution full fine-tuning with zero-calibration evaluation on held-out Pool~B users. The BiLSTM is reported separately as a conventional supervised reference and is not included in the following aggregate statistics over the nine foundation-model configurations. Across 171 model--dataset pairs, accuracy decreases in 141 cases (82.5\%) and macro-F1 in 146 cases (85.4\%). The average drop is 11.1 percentage points in accuracy and 14.9 points in macro-F1, with median drops of 7.5 and 11.2 points, respectively. As shown in Table~\ref{tab:rq3_generalization}, this degradation varies across models: MOMENT-S and MOMENT-B lose 27.1 and 22.4 macro-F1 points on average, respectively, while the two Chronos-2 variants lose approximately 19 points and UniTS loses 15.3 points. $\Delta_{\mathrm{New}}$ represents the overall gap between the seen-user Pool~A test split and the held-out Pool~B cohort, while LOSO provides a subject-disjoint evaluation within a fixed cohort. Strong in-distribution performance therefore does not guarantee strong performance on unseen users.

\vspace{-2mm}
\paragraph{Subject-wise variability.}
LOSO provides a complementary view by evaluating every subject in the selected cohort as a held-out test user. Across datasets, the median within-dataset LOSO standard deviation is 5.5 accuracy points, compared with a median 23.4-point range between model means. On CapgMyo, Pinch, and Mendeley Knee, the subject-level variability exceeds half of the between-model performance range. Individual configurations reach standard deviations of 21.7 points on Roshambo, 15.3 on Arm Translation, and 21.2 on Mendeley Knee. Thus, a single average cross-user score can conceal substantial variation across individual users.

\vspace{-2mm}
\paragraph{Limited calibration.}
We evaluate whether a small amount of labeled data from the new-user cohort can reduce this gap. Averaged over five seeds, five-shot adaptation improves macro-F1 in 120 of 171 model--dataset pairs (70.2\%). Relative to zero-calibration held-out evaluation, the average gain is 6.3 macro-F1 points, with a median gain of 4.8 points. The amount of recovery varies substantially across models: MOMENT-B improves on all 19 datasets, whereas UniTS improves on 13 of 19 datasets, and PatchTST and Lag-Llama decrease on average (Table~\ref{tab:rq3_generalization}). These results indicate that limited calibration is often helpful, but does not provide uniform recovery across tasks and users.

As a secondary observation, broad model ordering remains relatively stable under LOSO: the mean ranks under in-distribution FFT and LOSO have a Spearman correlation of $\rho=0.92$. This stability in relative ordering contrasts with the substantial drop in absolute performance and the large user-to-user variation observed above.

Taken together, performance decreases substantially when models are evaluated on new users, and average cross-user results can hide large differences across individual subjects. Limited five-shot adaptation improves performance in most settings, but the recovery is partial and uneven.
\vspace{-2mm}
\subsection{RQ4: Performance Consistency across EMG Tasks}
\label{sec:rq4}
\vspace{-2mm}
RQ4 examines whether model performance is consistent across different EMG tasks. We first compare upper- and lower-limb classification and then evaluate continuous EMG-to-text decoding on emg2qwerty. Table~\ref{tab:app_seq2seq} reports the Character Error Rate (CER, \%) on emg2qwerty.

Model performance is highly consistent between upper- and lower-limb classification, with Spearman correlations of $\rho=0.97$ for accuracy and $\rho=0.96$ for macro-F1. This consistency largely extends to continuous decoding: emg2qwerty performance correlates with upper-limb classification at $\rho = 0.85$ for both accuracy and macro-F1, and with lower-limb classification at $\rho = 0.78$ and $0.83$, respectively.

On emg2qwerty, Chronos-2-B achieves the lowest CER at 19.7\%, followed by Chronos-2-S at 24.0\%, MOMENT-B at 24.4\%, MOMENT-S at 26.8\%, and UniTS at 29.8\%. Lag-Llama obtains 38.0\%, while PatchTST, TimesFM, and BRANT range from 46.8\% to 54.9\%.

Overall, model performance remains strongly consistent across the evaluated task categories, with the strongest agreement between upper- and lower-limb classification. Because emg2qwerty is the only sequence-decoding dataset in the benchmark, additional datasets are needed to determine whether this consistency generalizes to other continuous decoding tasks.

\definecolor{headerblue}{RGB}{31, 56, 100}
\definecolor{rowlight}{RGB}{245, 245, 247}

\begin{table}[htbp]
\centering
\caption{Continuous typing decoding performance on the emg2qwerty dataset. All values represent Character Error Rate (CER, \%). Lower is better.
}
\label{tab:app_seq2seq}
\vspace{2mm}
\setlength{\tabcolsep}{3.2pt} 
\renewcommand{\arraystretch}{1.3} 

\rowcolors{2}{white}{rowlight} 

\resizebox{\textwidth}{!}{
\begin{tabular}{l ccc ccc ccc}
\rowcolor{headerblue} 
\color{white}\textbf{Dataset} &
\color{white}\textbf{MOMENT-S} &
\color{white}\textbf{MOMENT-B} &
\color{white}\textbf{BRANT} &
\color{white}\textbf{PatchTST} &
\color{white}\textbf{TimesFM} &
\color{white}\textbf{Chronos-2-S} &
\color{white}\textbf{Chronos-2-B} &
\color{white}\textbf{Lag-Llama} &
\color{white}\textbf{UniTS} \\ 
\midrule
emg2qwerty & 26.8 & 24.4 & 54.9 & 46.8 & 51.8 & 24.0 & \textbf{19.7} & 38.0 & 29.8 \\
\bottomrule
\end{tabular}
}
\end{table}

\vspace{-3mm}
\section{Conclusion}
\vspace{-2mm}
We introduce~\methodName, a systematic benchmark for diagnosing when and how pretrained time-series models transfer to EMG. Our results yield four practical lessons for future EMG foundation-model development. First, frozen evaluation is useful for identifying broad performance tiers but is insufficient for selecting a backbone intended for full adaptation; models should therefore be compared under the adaptation regime used at deployment. Second, strong fine-tuned performance is not itself evidence of useful pretraining. Architecture-matched scratch controls reveal that pretraining gains are concentrated in a small subset of configurations and do not increase consistently with model scale. Third, cross-user evaluation should report subject-level variability and macro-F1 in addition to average accuracy: absolute performance decreases substantially for unseen users, class imbalance can obscure failure cases, and limited calibration provides only partial and task-dependent recovery. Finally, model performance remains strongly consistent across the evaluated task categories, although agreement is stronger between upper- and lower-limb classification than with continuous EMG-to-text decoding.

\section*{AI Use Statement}
Generative AI tools were used to assist with literature search and
summarization, provide feedback on the presentation and interpretation of the methodology and experimental results, and improve the clarity, grammar, and organization of the manuscript. All cited works and AI-assisted summaries were independently verified against the original sources, and all AI-assisted text was reviewed and revised by the authors. The authors take full responsibility for the final content of this work.
\section*{Reproducibility Statement}
All 20 datasets in EMG-\textsc{FoundBench} are publicly available. Their sources, acquisition
hardware, sampling rates, channel configurations, and task definitions are documented in
Appendix~\ref{app:datasets} (Table~\ref{tab:datasets}). The unified preprocessing pipeline
(resampling, bandpass and notch filtering, instance normalization, and windowing) and the
Pool~A/Pool~B, LOSO, and five-shot partitioning procedures are described in
Appendix~\ref{app:preprocessing}. Model-specific input interfaces, pooling operations, and
classification and CTC output heads are specified in Appendix~\ref{app:classification_interfaces};
optimizer settings, learning rates, training budgets, batch sizes, checkpoint-selection rules,
and the five-shot random seeds are listed in Appendix~\ref{app:training}; the BiLSTM reference
architecture is described in Appendix~\ref{app:lstm}. Complete per-dataset results for every
evaluation protocol are reported in Appendix~\ref{app:complete_results}. Preprocessed datasets
in a unified format, preprocessing scripts, and evaluation code are released anonymously at
\url{https://anonymous.4open.science/r/EMG-FoundBench/} and
\url{https://huggingface.co/datasets/AnonymousAuthorSubmission/EMG-FoundBench}.

\bibliographystyle{unsrt}
\bibliography{ref}

\newpage
\appendix

\section{Appendix}

\label{app:complete_results}

This appendix provides complete per-dataset results supporting the four research questions examined in the main paper. RQ1 compares linear probing with full fine-tuning to evaluate frozen representation reuse. RQ2 compares pretrained full fine-tuning with architecture-matched training from scratch to assess the contribution of pretrained initialization. RQ3 reports zero-calibration held-out-user evaluation, cohort-restricted LOSO evaluation, and five-shot target-cohort adaptation. RQ4 examines the consistency of model behavior across upper-limb classification, lower-limb classification, and continuous EMG-to-text decoding. These complete results complement the aggregate analyses presented in Section~\ref{sec:results}.




\subsection{RQ1: Frozen Representation Reuse}

\label{app:id_performance}

RQ1 compares LP with FFT to determine whether EMG-relevant information is directly accessible from frozen pretrained representations and whether frozen-model performance predicts performance after full downstream adaptation. We first report the complete FFT results in Table~\ref{tab:app_downstream_all}. Models are trained on Pool A and evaluated on Pool A's held-out test split, providing an in-distribution reference under full downstream adaptation. All values are reported as accuracy and macro-F1 (\%).

\definecolor{headerblue}{RGB}{31, 56, 100}
\definecolor{rowlight}{RGB}{245, 245, 247}

\begin{table*}[htbp]
\centering
\caption{FFT performance across all 19 benchmark datasets. All values reported as Acc./F1 (\%).}
\label{tab:app_downstream_all}
\setlength{\tabcolsep}{3.2pt} 
\renewcommand{\arraystretch}{1.2} 
\rowcolors{2}{white}{rowlight} 

\resizebox{\textwidth}{!}{
\begin{tabular}{l ccc ccc ccc}
\rowcolor{headerblue} 
\color{white}\textbf{Dataset} &
\color{white}\textbf{MOMENT-S} &
\color{white}\textbf{MOMENT-B} &
\color{white}\textbf{BRANT} &
\color{white}\textbf{PatchTST} &
\color{white}\textbf{TimesFM} &
\color{white}\textbf{Chronos-2-S} &
\color{white}\textbf{Chronos-2-B} &
\color{white}\textbf{Lag-Llama} &
\color{white}\textbf{UniTS} \\ 
\midrule

NinaPro DB1 & 43.6/48.1 & 32.5/35.8 & 22.0/24.1 & 22.2/25.5 & 38.9/39.6 & 42.5/46.6 & \textbf{44.8/49.0} & 17.5/12.5 & 36.1/39.0 \\
NinaPro DB2 & 55.4/55.2 & 26.1/25.6 & 3.5/1.0 & 6.8/6.2 & 17.4/16.4 & \textbf{56.6/56.7} & 49.2/49.3 & 2.4/0.2 & 12.8/11.6 \\
NinaPro DB3 & \textbf{28.1/28.1} & 13.0/12.0 & 3.7/0.6 & 2.1/2.1 & 3.7/2.3 & 22.9/22.0 & 21.7/20.9 & 2.7/0.4 & 5.6/3.2 \\
NinaPro DB4 & \textbf{44.3/45.5} & 23.7/22.5 & 4.0/0.5 & 9.5/8.2 & 9.2/8.0 & 34.1/35.5 & 40.5/43.1 & 4.5/0.4 & 12.6/11.5 \\
NinaPro DB5 & \textbf{28.6/29.0} & 11.7/10.8 & 6.7/1.0 & 6.1/5.3 & 10.5/8.7 & 23.0/23.3 & 22.2/22.5 & 6.4/1.2 & 10.8/7.8 \\
NinaPro DB6 & 51.0/53.1 & 38.2/38.3 & 15.7/9.9 & 21.4/21.4 & 43.0/32.4 & \textbf{58.9}/59.1 & 58.3/\textbf{59.9} & 56.4/58.2 & 23.0/19.6 \\ 
NinaPro DB7 & \textbf{66.4/67.1} & 40.5/40.5 & 3.4/0.6 & 9.9/10.0 & 24.9/25.2 & 60.4/60.2 & 55.7/55.9 & 5.8/0.5 & 27.2/27.0 \\
GRABMyo & \textbf{74.0/74.0} & 70.6/70.9 & 12.7/8.6 & 19.5/18.3 & 19.8/17.9 & 54.3/54.3 & 62.1/61.7 & 34.4/28.8 & 63.8/63.6 \\ 
CapgMyo & \textbf{88.2/88.1} & 66.0/65.1 & 11.8/4.2 & 20.4/10.3 & 16.0/13.0 & 25.0/25.1 & 21.5/21.5 & 53.3/40.0 & 22.2/17.4 \\
Pinch & \textbf{60.8}/45.7 & 55.4/\textbf{47.4} & 49.5/13.2 & 38.7/23.0 & 47.4/46.0 & 48.9/33.4 & 54.3/38.4 & 45.7/13.8 & 49.5/25.7 \\
Roshambo & \textbf{76.9/78.3} & 38.5/32.2 & 38.5/30.3 & 49.2/48.9 & 58.7/47.2 & 41.5/41.8 & 64.6/63.9 & 41.5/19.6 & 50.8/49.8 \\
emg2pose & 32.5/10.2 & 30.8/10.0 & 25.1/2.9 & 38.9/21.2 & 36.7/16.1 & \textbf{77.8/73.6} & 69.1/61.3 & 53.8/45.7 & 20.0/19.7 \\ 
3DC & 56.4/49.5 & 63.7/55.3 & 18.7/13.6 & 34.1/29.9 & 36.4/27.4 & 54.2/48.1 & 59.0/52.4 & \textbf{83.9/77.4} & 36.3/29.5 \\
ContinuousTrans & 95.5/95.5 & 78.6/78.5 & 27.6/23.8 & 68.4/68.5 & 80.0/80.1 & \textbf{97.2/97.2} & \textbf{97.2/97.2} & 93.9/94.0 & 89.2/89.2 \\
Arm Translation & \textbf{94.8/94.1} & 83.8/82.6 & 42.4/29.6 & 61.9/57.0 & 69.4/64.6 & 91.1/90.1 & 90.5/89.2 & 74.1/69.6 & 82.7/81.1 \\
\midrule
HD-EMG & \textbf{71.8/71.0} & 66.4/65.7 & 48.0/36.2 & 43.1/40.9 & 36.3/34.5 & 47.5/43.7 & 46.2/46.1 & 39.5/37.1 & 56.1/53.2 \\
SIAT-LLMD & \textbf{49.9/55.0} & 37.1/39.0 & 16.5/15.3 & 23.2/29.6 & 31.6/36.0 & 44.8/50.6 & 45.7/51.9 & 23.6/24.1 & 35.6/40.7 \\
Mendeley Knee & 67.7/67.6 & 63.8/63.8 & 50.0/38.3 & 51.7/51.6 & 57.3/57.0 & 70.9/70.9 & \textbf{95.9/95.9} & 62.4/62.4 & 75.7/75.2 \\
MyPredict & 56.9/44.8 & 49.2/41.2 & 33.7/8.8 & 44.2/18.1 & \textbf{83.3/69.8} & 80.3/58.9 & 74.5/56.5 & 20.9/6.2 & 53.8/26.9 \\


\bottomrule
\end{tabular}
}
\end{table*}

\subsubsection{Linear Probing versus Full Fine-Tuning}
\label{app:lp_vs_fft}

Tables~\ref{tab:app_downstream_all} and~\ref{tab:app_full_finetune_all} report the complete FFT and LP results across all 19 classification datasets, respectively. Under LP, the pretrained backbone is frozen and only the classification head is trained, whereas FFT updates both the backbone and the classification head. Comparing these two adaptation regimes evaluates whether EMG-relevant information is directly accessible from frozen pretrained representations and whether model performance under LP predicts performance after full downstream adaptation.

\begin{table*}[htbp]
\centering
\caption{Complete Linear Probe Performance across all 19 datasets. All values reported as Acc./F1 (\%).}
\label{tab:app_full_finetune_all}
\setlength{\tabcolsep}{2.8pt} 
\renewcommand{\arraystretch}{1.2} 
\rowcolors{2}{white}{rowlight} 

\resizebox{\textwidth}{!}{
\begin{tabular}{l ccc ccc ccc}
\rowcolor{headerblue} 
\color{white}\textbf{Dataset} &
\color{white}\textbf{MOMENT-S} &
\color{white}\textbf{MOMENT-B} &
\color{white}\textbf{BRANT} &
\color{white}\textbf{PatchTST} &
\color{white}\textbf{TimesFM} &
\color{white}\textbf{Chronos-2-S} &
\color{white}\textbf{Chronos-2-B} &
\color{white}\textbf{Lag-Llama} &
\color{white}\textbf{UniTS} \\ 
\midrule
NinaPro DB1 & 26.5/26.8 & \textbf{32.0/35.5} & 11.9/8.9 & 9.7/7.8 & 6.3/4.3 & 16.6/18.1 & 16.8/17.7 & 15.7/11.3 & 10.8/5.5 \\
NinaPro DB2 & 20.5/20.6 & \textbf{33.0/31.9} & 0.9/0.0 & 2.3/1.0 & 5.7/5.1 & 8.2/6.2 & 8.2/6.4 & 1.7/0.1 & 2.5/0.8 \\
NinaPro DB3 & 8.7/8.9 & \textbf{19.6/19.0} & 0.2/0.0 & 2.3/0.8 & 2.3/1.1 & 3.5/2.0 & 3.9/2.0 & 2.9/0.5 & 2.5/1.3 \\
NinaPro DB4 & 23.0/19.5 & \textbf{33.4/34.6} & 5.2/0.8 & 5.5/4.0 & 5.7/2.9 & 7.1/3.2 & 10.4/5.8 & 4.3/0.7 & 3.8/1.2 \\
NinaPro DB5 & 14.9/10.2 & \textbf{22.2/21.5} & 7.0/1.1 & 5.5/1.1 & 7.0/3.1 & 8.5/3.7 & 8.2/3.7 & 5.3/1.1 & 6.4/1.1 \\
NinaPro DB6 & 36.3/36.6 & \textbf{44.4/46.3} & 5.0/1.4 & 6.8/3.9 & 35.9/31.0 & 19.0/17.7 & 21.3/19.6 & 12.9/6.6 & 35.3/30.7 \\ 
NinaPro DB7 & 35.6/32.3 & \textbf{52.1/52.5} & 5.1/0.3 & 5.7/2.3 & 7.9/7.6 & 13.0/8.4 & 14.4/10.3 & 3.4/0.6 & 4.3/1.1 \\
GRABMyo & 57.4/56.4 & \textbf{70.3/69.9} & 6.8/3.5 & 10.2/7.0 & 20.6/20.6 & 20.6/17.2 & 22.8/18.6 & 30.7/28.4 & 61.7/61.8 \\ 
CapgMyo & 45.4/23.4 & \textbf{77.1/75.8} & 9.7/3.1 & 36.1/18.9 & 23.6/23.0 & 13.2/12.2 & 22.2/19.0 & 33.0/7.6 & 11.1/8.6 \\
Pinch & 48.9/13.2 & \textbf{54.3/46.9} & 49.5/13.2 & 27.4/19.4 & 39.8/39.2 & 47.3/18.4 & 48.9/13.1 & 26.3/12.4 & 49.5/13.2 \\
Roshambo & 53.8/51.7 & \textbf{69.2/68.7} & 27.7/14.5 & 30.8/26.3 & 65.1/64.8 & 40.0/33.6 & 38.5/28.8 & 30.8/25.6 & 40.0/31.1 \\
emg2pose & 36.3/17.2 & 34.2/\textbf{18.9} & 24.3/2.5 & 25.4/2.9 & 33.9/9.8 & 36.6/14.5 & \textbf{37.8}/16.4 & 31.0/17.8 & 27.5/15.2 \\ 
3DC & 29.3/23.1 & \textbf{41.4/35.6} & 2.9/1.0 & 5.5/2.9 & 23.4/28.6 & 24.9/17.8 & 24.5/19.9 & 8.8/3.2 & 8.8/4.5 \\
ContinuousTrans & 79.4/79.3 & \textbf{83.8/83.7} & 17.8/7.3 & 22.4/17.0 & 53.0/38.1 & 59.9/58.6 & 62.8/61.7 & 58.0/54.9 & 28.9/24.9 \\
Arm Translation & 82.7/80.6 & \textbf{87.0/86.4} & 14.3/4.2 & 38.3/26.2 & 54.8/43.3 & 57.8/50.7 & 58.6/51.0 & 12.7/3.8 & 35.6/23.0 \\
\midrule
HD-EMG & 58.7/57.7 & \textbf{78.5/78.3} & 44.4/20.5 & 35.4/29.1 & 30.4/29.1 & 42.6/25.3 & 44.8/31.0 & 44.0/20.4 & 45.3/22.5 \\
SIAT-LLMD & 21.6/22.9 & \textbf{31.6/35.7} & 3.2/1.0 & 6.9/7.1 & 18.5/20.7 & 13.1/16.7 & 15.7/19.9 & 9.6/9.8 & 12.6/11.3 \\
Mendeley Knee & 63.4/63.3 & \textbf{66.3/66.3} & 53.7/41.6 & 51.4/51.4 & 54.6/54.5 & 62.1/62.1 & 59.4/59.3 & 49.8/49.5 & 50.5/50.1 \\
MyPredict & 54.1/23.0 & 61.0/31.0 & 55.7/11.5 & 56.5/11.4 & \textbf{70.7/37.7} & 39.4/11.0 & 38.8/11.6 & 27.7/6.6 & 61.0/11.8 \\

\bottomrule
\end{tabular}
}
\end{table*}

\subsection{RQ2: Contribution of Pretrained Initialization}
\label{app:pretraining_contribution}

RQ2 compares the pretrained FFT results in Table~\ref{tab:app_downstream_all} with architecture-matched train-from-scratch results. For each model, the two settings use the same backbone architecture, input adapter, pooling operation, classification head, data splits, and checkpoint-selection procedure, while differing in backbone initialization. The pretrained setting initializes the backbone from the released checkpoint, whereas the scratch setting randomly initializes the same backbone. Table~\ref{tab:app_scratch_all} reports the complete train-from-scratch results across all 19 classification datasets. Together, these results assess the contribution of pretrained initialization beyond the underlying architecture under the corresponding downstream training settings. In addition to the architecture-matched scratch controls used in RQ2, we include a medium-capacity bidirectional LSTM as a conventional non-pretrained temporal reference. Complete architectural and training details are provided in Appendix~\ref{app:lstm}.

\begin{table*}[htbp]
\centering
\caption{Complete train-from-scratch performance across all benchmark datasets. 
Models are randomly initialized and trained using the same downstream full-training protocol as the pretrained full fine-tuning experiments. 
All values are reported as Acc./F1 (\%).}
\label{tab:app_scratch_all}

\setlength{\tabcolsep}{2.8pt} 
\renewcommand{\arraystretch}{1.2} 
\rowcolors{2}{white}{rowlight} 

\resizebox{\textwidth}{!}{
\begin{tabular}{l ccc ccc ccc c}
\rowcolor{headerblue} 
\color{white}\textbf{Dataset} &
\color{white}\textbf{MOMENT-S} &
\color{white}\textbf{MOMENT-B} &
\color{white}\textbf{BRANT} &
\color{white}\textbf{PatchTST} &
\color{white}\textbf{TimesFM} &
\color{white}\textbf{Chronos-2-S} &
\color{white}\textbf{Chronos-2-B} &
\color{white}\textbf{Lag-Llama} &
\color{white}\textbf{UniTS} &
\color{white}\textbf{BiLSTM} \\ 
\midrule

NinaPro DB1 & 41.1/45.1 & 41.6/45.4 & 20.0/19.3 & 23.3/25.6 & \textbf{47.4/50.7} & 25.1/29.0 & 29.0/32.3 & 24.3/27.0 & 28.6/32.1 & 38.9/41.7 \\
NinaPro DB2 & 44.6/45.4 & \textbf{52.4/52.9} & 1.5/0.4 & 8.9/8.4 & 8.4/8.0 & 7.5/5.4 & 17.3/16.5 & 25.1/25.4 & 8.9/8.2 & 9.4/9.7 \\
NinaPro DB3 & \textbf{21.5/20.7} & 19.5/18.8 & 1.7/0.6 & 2.6/2.5 & 2.9/2.4 & 4.3/1.2 & 4.7/3.1 & 6.0/6.2 & 3.5/1.9 & 2.6/2.4 \\
NinaPro DB4 & \textbf{38.7/39.4} & 36.3/35.0 & 5.7/1.2 & 8.2/7.8 & 4.5/3.6 & 10.0/8.5 & 20.9/19.7 & 13.7/14.3 & 9.0/5.7 & 7.7/7.1 \\
NinaPro DB5 & \textbf{30.7/31.0} & 22.7/25.4 & 7.3/1.4 & 9.9/8.4 & 6.7/4.1 & 9.3/6.1 & 14.3/10.6 & 12.0/12.1 & 6.4/0.6 & 5.3/4.4 \\
NinaPro DB6 & 46.7/48.0 & \textbf{46.8/48.8} & 7.1/4.4 & 25.9/25.4 & 23.9/24.0 & 33.2/34.5 & 38.8/40.4 & 35.7/36.1 & 43.2/44.7 & 17.8/17.5 \\
NinaPro DB7 & 62.4/62.1 & \textbf{67.1/68.3} & 2.5/0.5 & 10.6/8.9 & 5.8/4.6 & 13.5/9.2 & 18.7/18.5 & 44.3/44.2 & 18.2/17.5 & 12.4/13.4 \\
GRABMyo & 77.7/77.5 & 78.9/78.7 & 9.3/3.8 & 30.5/29.1 & 19.5/18.7 & 39.2/36.9 & 38.1/37.4 & 62.9/63.0 & \textbf{79.8/79.9} & 39.3/39.0 \\
CapgMyo & \textbf{76.9/76.7} & 71.3/71.1 & 15.3/8.5 & 22.5/19.2 & 16.0/15.0 & 23.6/21.3 & 22.2/19.7 & 49.3/49.3 & 16.0/11.1 & 13.1/13.7 \\
Pinch & 62.7/\textbf{51.8} & \textbf{63.2}/50.7 & 49.5/13.2 & 45.5/23.3 & 41.4/27.4 & 52.7/25.9 & 51.6/26.6 & 41.4/23.2 & 56.5/43.6 & 38.8/27.7 \\
Roshambo & 65.3/64.6 & \textbf{70.8/72.3} & 35.4/29.7 & 43.1/40.2 & 41.5/41.9 & 32.3/32.5 & 36.9/35.4 & 53.9/53.4 & 27.7/14.5 & 68.1/68.9 \\
emg2pose & 51.5/44.4 & 50.1/38.5 & 19.4/11.2 & 33.4/14.9 & 50.4/38.4 & \textbf{66.6/58.8} & 64.5/55.0 & 39.1/25.3 & 51.8/41.4 & 32.2/17.6 \\
3DC & 59.0/55.6 & \textbf{59.3/57.7} & 8.4/5.7 & 32.6/32.3 & 26.7/22.8 & 39.6/27.3 & 38.8/35.1 & 28.2/23.8 & 28.2/20.3 & 19.6/15.7 \\
ContinuousTrans & \textbf{95.0/95.0} & 90.4/90.3 & 24.7/18.9 & 72.6/72.2 & 69.9/69.5 & 92.3/92.4 & 92.5/92.5 & 84.1/84.0 & 87.9/87.9 & 74.5/74.1 \\
Arm Translation & 89.9/89.1 & \textbf{91.0/89.9} & 42.5/27.6 & 58.9/52.8 & 57.6/51.8 & 69.0/65.0 & 79.8/77.4 & 86.4/84.6 & 78.6/76.7 & 61.3/56.7 \\
\midrule
HD-EMG & 62.5/61.8 & \textbf{66.1/63.5} & 38.1/28.9 & 37.1/33.2 & 46.2/43.3 & 52.9/48.0 & 42.2/37.7 & 53.4/52.0 & 44.4/20.5 & 42.7/42.0 \\
SIAT-LLMD & \textbf{50.1/54.5} & 42.1/45.1 & 18.3/17.0 & 24.7/31.6 & 28.4/36.1 & 25.7/28.7 & 31.9/39.1 & 35.0/39.1 & 28.0/33.7 & 43.6/47.6 \\
Mendeley Knee & 62.1/62.0 & \textbf{76.5/76.5} & 47.8/45.2 & 55.2/55.2 & 55.1/54.5 & 65.0/63.6 & 60.0/59.8 & 65.7/65.7 & 55.4/55.3 & 64.2/64.2 \\
MyPredict & 52.2/40.9 & 43.0/17.6 & 41.3/32.1 & 37.6/18.8 & 48.7/31.0 & 64.2/42.2 & \textbf{66.0}/41.2 & 49.4/31.8 & 49.4/23.5 & 64.0/\textbf{45.8} \\

\bottomrule
\end{tabular}
}
\end{table*}

\subsection{RQ3: Cross-User Transfer and Calibration}

\label{app:cross_user}

\subsubsection{Zero-Calibration Held-Out-User Evaluation}

\label{app:zero_shot}

Table~\ref{tab:app_zero_shot_all} reports complete zero-calibration results across all 19 classification datasets. Models trained on Pool A, comprising approximately 90\% of the subjects, are evaluated directly on the held-out Pool B subjects without target-user adaptation. This protocol measures how well the Pool A-trained models generalize to previously unseen users while keeping the trained model fixed. All values are reported as accuracy and macro-F1 (\%).

\begin{table*}[htbp]
\centering
\caption{Complete zero-calibration cross-subject generalization performance across all 19 benchmark classification datasets. Models trained on Pool A, evaluated on Pool B without adaptation. All values reported as Acc./F1 (\%).}
\label{tab:app_zero_shot_all}

\setlength{\tabcolsep}{3.2pt} 
\renewcommand{\arraystretch}{1.3} 

\rowcolors{2}{white}{rowlight} 

\resizebox{\textwidth}{!}{
\begin{tabular}{l ccc ccc ccc c}
\rowcolor{headerblue} 
\color{white}\textbf{Dataset} &
\color{white}\textbf{MOMENT-S} &
\color{white}\textbf{MOMENT-B} &
\color{white}\textbf{BRANT} &
\color{white}\textbf{PatchTST} &
\color{white}\textbf{TimesFM} &
\color{white}\textbf{Chronos-2-S} &
\color{white}\textbf{Chronos-2-B} &
\color{white}\textbf{Lag-Llama} &
\color{white}\textbf{UniTS} &
\color{white}\textbf{BiLSTM} \\ 
\midrule
NinaPro DB1 & 23.8/25.0 & 25.4/22.4 & 12.4/13.7 & 21.2/19.9 & \textbf{43.0/41.8} & 21.2/25.2 & 28.0/27.1 & 16.1/9.6 & 22.8/22.7 & 17.6/15.2 \\
NinaPro DB2 & 26.9/23.9 & 18.9/17.2 & 1.5/0.4 & 8.0/4.4 & 13.4/9.7 & \textbf{27.9/25.8} & 25.4/24.0 & 3.5/0.3 & 12.1/8.5 & 11.4/10.7 \\
NinaPro DB3 & 7.8/5.4 & 1.9/1.4 & 1.0/0.0 & 1.9/1.1 & 7.8/2.8 & 7.8/4.1 & \textbf{13.6/11.6} & 1.9/0.1 & 5.8/1.9 & 1.0/1.0 \\
NinaPro DB4 & \textbf{37.2}/\textbf{30.2} & 9.3/6.6 & 2.3/0.2 & 7.0/3.7 & 4.7/4.6 & 27.9/20.3 & \textbf{37.2}/28.1 & 2.3/0.2 & 14.0/11.9 & 11.6/10.5 \\
NinaPro DB5 & 25.0/\textbf{26.0} & 9.4/4.3 & 6.3/0.7 & 6.3/3.2 & 12.5/7.4 & \textbf{31.3}/23.9 & 15.6/8.6 & 3.1/1.2 & 6.2/2.6 & 6.3/3.9 \\
NinaPro DB6 & 38.0/29.1 & 29.5/27.4 & 5.5/1.5 & 19.0/15.8 & 23.9/22.4 & 42.3/37.5 & \textbf{48.5/41.7} & 35.6/25.8 & 13.4/11.1 & 26.4/23.7 \\
NinaPro DB7 & \textbf{54.6/47.2} & 35.2/32.0 & 1.2/0.6 & 6.8/4.7 & 33.0/26.7 & 44.3/34.9 & 37.5/27.4 & 2.3/0.1 & 25.0/17.3 & 12.5/9.2 \\
GRABMyo & \textbf{60.0/56.9} & 50.6/52.5 & 8.2/3.3 & 21.2/20.0 & 17.7/16.7 & 47.1/47.1 & 49.4/49.9 & 36.5/32.0 & 10.3/6.9 & 34.1/31.0 \\
CapgMyo & 25.0/18.9 & 12.5/4.2 & 6.3/1.7 & \textbf{33.3}/11.4 & 12.5/8.6 & 21.5/\textbf{21.5} & 6.3/5.0 & 18.8/8.3 & 18.8/11.9 & 6.3/3.1 \\
Pinch & 47.8/23.3 & 34.8/15.7 & \textbf{60.9}/15.1 & 34.8/11.9 & 43.5/24.1 & 39.1/23.9 & 52.2/36.1 & \textbf{60.9}/15.1 & 47.8/25.0 & 56.5/\textbf{43.3} \\
Roshambo & 28.6/22.9 & 42.9/35.6 & \textbf{57.1}/45.7 & \textbf{57.1}/44.4 & 42.9/20.0 & 28.6/14.8 & \textbf{57.1}/\textbf{57.8} & 42.9/20.0 & 42.9/32.4 & 42.9/38.9 \\
emg2pose & 19.6/12.3 & 29.6/8.7 & 25.1/3.0 & 38.6/22.0 & 34.9/15.0 & \textbf{67.7/62.4} & 65.2/57.4 & 50.4/37.1 & 22.8/9.0 & 28.9/13.9 \\
3DC & 20.4/13.5 & 16.3/12.8 & 10.2/1.7 & 22.5/17.8 & 14.3/11.5 & 22.5/19.5 & 32.7/27.1 & \textbf{34.7/28.0} & 14.3/8.0 & 12.2/7.0 \\
ContinuousTrans & 73.7/69.1 & 59.2/57.0 & 24.2/17.4 & 54.3/52.1 & 63.0/62.1 & 81.0/79.1 & \textbf{85.8/85.4} & 72.0/70.4 & 69.6/66.7 & 56.4/55.2 \\
Arm Translation & \textbf{65.6}/51.8 & 20.0/16.9 & 46.7/33.8 & 47.8/35.4 & 50.0/39.2 & 60.0/44.7 & 64.4/\textbf{52.6} & 60.0/44.3 & 42.2/34.9 & 30.0/28.5 \\
\midrule
HD-EMG & 48.0/39.8 & 40.0/30.4 & 36.0/22.2 & 44.0/43.2 & \textbf{60.0/55.5} & 44.0/39.3 & 52.0/45.5 & 36.0/21.4 & 40.0/37.1 & 32.0/30.1 \\
SIAT-LLMD & \textbf{52.9/38.1} & 36.5/29.7 & 11.8/10.6 & 4.6/2.7 & 28.6/14.7 & 44.9/32.5 & 48.7/33.5 & 33.1/27.8 & 44.9/30.1 & 45.2/35.0 \\
Mendeley Knee & 38.1/38.1 & 26.2/25.8 & 33.3/30.1 & 50.0/50.0 & 44.1/43.9 & 52.4/52.1 & 41.7/40.2 & 19.1/16.0 & 48.8/47.9 & \textbf{53.5/53.3} \\
MyPredict & 48.6/13.9 & 36.6/11.8 & 26.2/12.9 & 38.0/11.9 & 40.5/17.1 & \textbf{79.3/19.3} & 71.1/17.3 & 11.0/7.1 & 25.3/14.5 & 55.9/14.5 \\


\bottomrule
\end{tabular}
}
\end{table*}

\subsubsection{Cohort-Restricted LOSO Evaluation}

Table~\ref{tab:app_loso_all} reports the complete cohort-restricted LOSO results across all 19 classification datasets. Datasets containing at most 10 subjects use all available subjects, whereas datasets containing more than 10 subjects use a fixed cohort consisting of the first 10 subjects according to the canonical subject-ID ordering of the original dataset. The same cohort is used for every model configuration. Within each cohort, every subject is held out once for testing while the remaining cohort members are used for downstream training and validation. This protocol limits evaluation to at most 10 folds per dataset while maintaining strict subject separation between training and testing. Results are reported as mean accuracy $\pm$ standard deviation across the held-out subjects within each evaluation cohort. The fixed cohort is used as a computationally standardized evaluation subset rather than as a population-representative sample.

\begin{table*}[htbp]
\centering
\caption{Complete cohort-restricted leave-one-subject-out (LOSO) results across all 19 classification datasets. For each dataset, one subject from the evaluation cohort is held out for testing while the remaining cohort members are used for downstream training and validation. Datasets with at most 10 subjects use all available subjects, whereas larger datasets use a fixed 10-subject cohort. Results are reported as mean accuracy $\pm$ standard deviation (\%) across held-out subjects.}
\label{tab:app_loso_all}

\setlength{\tabcolsep}{2.8pt}
\renewcommand{\arraystretch}{1.2}
\rowcolors{2}{white}{rowlight}

\resizebox{\textwidth}{!}{
\begin{tabular}{l ccc ccc ccc c}
\rowcolor{headerblue}
\color{white}\textbf{Dataset} &
\color{white}\textbf{MOMENT-S} &
\color{white}\textbf{MOMENT-B} &
\color{white}\textbf{BRANT} &
\color{white}\textbf{PatchTST} &
\color{white}\textbf{TimesFM} &
\color{white}\textbf{Chronos-2-S} &
\color{white}\textbf{Chronos-2-B} &
\color{white}\textbf{Lag-Llama} &
\color{white}\textbf{UniTS} &
\color{white}\textbf{BiLSTM} \\
\midrule

NinaPro DB1 & $16.7\pm3.8$ & $19.0\pm4.9$ & $7.9\pm1.5$ & $6.8\pm0.9$ & $\mathbf{40.6\pm5.3}$ & $19.9\pm3.5$ & $21.3\pm3.3$ & $10.7\pm2.0$ & $13.2\pm3.2$ & $8.5\pm3.1$ \\
NinaPro DB2 & $17.5\pm3.5$ & $\mathbf{18.1\pm3.3}$ & $7.6\pm1.5$ & $5.7\pm0.5$ & $9.3\pm2.6$ & $14.7\pm4.2$ & $15.0\pm3.6$ & $16.8\pm4.7$ & $12.7\pm3.1$ & $10.7\pm4.9$ \\
NinaPro DB3 & $\mathbf{5.9\pm2.5}$ & $5.4\pm3.0$ & $5.8\pm1.3$ & $5.3\pm1.2$ & $3.1\pm1.4$ & $2.8\pm1.3$ & $3.5\pm1.5$ & $5.3\pm2.5$ & $5.7\pm2.1$ & $5.6\pm1.1$ \\
NinaPro DB4 & $19.2\pm3.3$ & $18.9\pm3.0$ & $5.7\pm0.8$ & $6.2\pm1.3$ & $13.6\pm3.5$ & $23.0\pm4.2$ & $\mathbf{24.7\pm5.5}$ & $14.6\pm2.4$ & $15.5\pm2.2$ & $5.9\pm1.3$ \\
NinaPro DB5 & $14.5\pm3.4$ & $13.8\pm3.7$ & $6.6\pm0.8$ & $6.5\pm0.9$ & $10.8\pm1.9$ & $16.2\pm2.9$ & $\mathbf{18.0\pm3.2}$ & $9.6\pm3.1$ & $12.1\pm2.5$ & $5.6\pm1.5$ \\
NinaPro DB6 & $34.3\pm5.6$ & $33.7\pm4.1$ & $17.0\pm1.8$ & $15.3\pm2.0$ & $37.1\pm3.0$ & $42.9\pm7.1$ & $\mathbf{46.5\pm6.7}$ & $28.6\pm3.0$ & $29.1\pm2.4$ & $18.3\pm1.7$ \\
NinaPro DB7 & $25.6\pm7.4$ & $\mathbf{27.1\pm10.5}$ & $4.5\pm1.3$ & $3.7\pm1.2$ & $14.5\pm3.1$ & $26.9\pm4.7$ & $26.6\pm8.6$ & $19.9\pm4.6$ & $20.2\pm4.2$ & $6.8\pm2.4$ \\
GRABMyo & $25.8\pm7.5$ & $\mathbf{30.1\pm8.7}$ & $6.0\pm0.3$ & $7.5\pm2.5$ & $22.9\pm4.4$ & $19.9\pm6.0$ & $21.1\pm5.8$ & $27.7\pm6.3$ & $27.1\pm5.5$ & $11.5\pm4.0$ \\
CapgMyo & $\mathbf{22.0\pm11.4}$ & $16.0\pm6.3$ & $12.6\pm0.9$ & $12.3\pm3.9$ & $21.0\pm5.6$ & $16.8\pm4.3$ & $14.3\pm6.9$ & $18.5\pm5.5$ & $17.9\pm4.9$ & $13.1\pm3.0$ \\
Pinch & $46.9\pm4.8$ & $45.8\pm10.0$ & $\mathbf{56.1\pm14.5}$ & $38.0\pm9.0$ & $41.2\pm8.2$ & $46.4\pm9.4$ & $50.3\pm10.3$ & $47.0\pm6.7$ & $47.7\pm10.7$ & $44.8\pm6.0$ \\
Roshambo & $\mathbf{64.5\pm21.7}$ & $62.0\pm18.7$ & $31.1\pm7.6$ & $27.5\pm8.9$ & $38.4\pm9.7$ & $37.8\pm7.5$ & $43.5\pm8.5$ & $54.8\pm9.5$ & $47.3\pm15.1$ & $57.5\pm17.5$ \\
emg2pose & $30.4\pm3.0$ & $30.2\pm2.9$ & $33.1\pm1.2$ & $24.8\pm3.3$ & $31.5\pm2.3$ & $\mathbf{41.0\pm4.7}$ & $39.9\pm4.0$ & $30.0\pm5.0$ & $30.9\pm3.4$ & $31.6\pm2.9$ \\
3DC & $13.5\pm4.7$ & $20.4\pm6.9$ & $9.1\pm0.6$ & $9.7\pm2.4$ & $29.7\pm7.4$ & $30.5\pm9.1$ & $\mathbf{37.5\pm7.8}$ & $24.7\pm6.5$ & $18.0\pm4.6$ & $12.4\pm4.6$ \\
ContinuousTrans & $65.0\pm12.0$ & $64.5\pm10.7$ & $27.5\pm7.3$ & $20.8\pm2.9$ & $67.0\pm8.9$ & $69.0\pm16.8$ & $\mathbf{71.2\pm15.6}$ & $55.6\pm16.6$ & $55.0\pm11.3$ & $50.9\pm13.1$ \\
Arm Translation & $54.3\pm11.6$ & $57.5\pm14.8$ & $36.3\pm3.4$ & $29.2\pm3.2$ & $54.4\pm9.7$ & $63.0\pm10.5$ & $\mathbf{64.2\pm8.8}$ & $50.8\pm15.3$ & $48.5\pm9.2$ & $54.6\pm11.0$ \\

\midrule

HD-EMG & $50.9\pm8.0$ & $50.1\pm11.2$ & $38.7\pm14.0$ & $43.1\pm2.1$ & $\mathbf{57.4\pm4.2}$ & $50.9\pm7.7$ & $54.6\pm8.9$ & $48.3\pm5.4$ & $53.8\pm8.5$ & $43.2\pm3.1$ \\
SIAT-LLMD & $27.6\pm3.3$ & $32.2\pm4.0$ & $10.1\pm0.9$ & $9.5\pm0.8$ & $25.2\pm3.0$ & $32.7\pm4.1$ & $\mathbf{37.0\pm3.7}$ & $19.0\pm2.1$ & $27.9\pm2.0$ & $27.3\pm1.9$ \\
Mendeley Knee & $60.9\pm17.1$ & $63.7\pm19.5$ & $61.6\pm11.8$ & $53.3\pm4.9$ & $\mathbf{66.7\pm16.1}$ & $62.3\pm21.2$ & $ 59.0\pm16.8 $ & $51.0\pm7.1$ & $53.6\pm10.1$ & $52.6\pm6.7$ \\
MyPredict & $67.6\pm8.1$ & $67.3\pm8.7$ & $59.5\pm7.3$ & $46.9\pm5.2$ & $66.0\pm8.5$ & $80.7\pm8.3$ & $\mathbf{82.6\pm6.9}$ & $56.1\pm11.0$ & $55.9\pm16.7$ & $72.3\pm6.9$ \\

\bottomrule
\end{tabular}
}
\end{table*}

\subsubsection{Five-Shot Target-Cohort Adaptation}

\label{app:few_shot}

Table~\ref{tab:app_few_shot_all} reports complete five-shot adaptation results across all 19 classification datasets. Starting from models trained on Pool~A, we randomly sample five labeled examples per class from the pooled Pool~B training split for adaptation and evaluate on the fixed Pool~B test split. The five examples are sampled per class from the pooled Pool~B adaptation split, rather than separately for each target user. We repeat this procedure with five random seeds (42--46), with the Pool~A--Pool~B subject partition and test split held fixed. Each seed controls both the sampled adaptation examples and the adaptation run, so the same seed-specific adaptation samples are used across all model configurations. Comparing these results with the zero-calibration results in Table~\ref{tab:app_zero_shot_all} quantifies how much cross-subject performance can be recovered using limited supervision from the target cohort. Results are reported as the mean and standard deviation of macro-F1 (\%) across the five seeds.

\begin{table*}[htbp]
\centering
\caption{Complete five-shot target-cohort adaptation results across all 19 classification datasets (macro-F1). For each of five random seeds, models are adapted using five labeled examples per class sampled from the pooled Pool B training set and evaluated on the fixed Pool B test set. Values are reported as mean $\pm$ standard deviation of macro-F1 (\%) across 5 random seeds (42--46).}
\label{tab:app_few_shot_all}
\setlength{\tabcolsep}{3.2pt}
\renewcommand{\arraystretch}{1.3}

\rowcolors{2}{white}{rowlight}

\resizebox{\textwidth}{!}{
\begin{tabular}{l ccc c ccc c c c}
\rowcolor{headerblue}
\color{white}\textbf{Dataset} &
\color{white}\textbf{MOMENT-S} &
\color{white}\textbf{MOMENT-B} &
\color{white}\textbf{BRANT} &
\color{white}\textbf{PatchTST} &
\color{white}\textbf{TimesFM} &
\color{white}\textbf{Chronos-2-S} &
\color{white}\textbf{Chronos-2-B} &
\color{white}\textbf{Lag-Llama} &
\color{white}\textbf{UniTS} &
\color{white}\textbf{BiLSTM} \\
\midrule
NinaPro DB1      & $\mathbf{44.0{\pm}9.9}$ & $29.0{\pm}3.7$ & $6.9{\pm}2.6$ & $6.1{\pm}0.9$ & $32.5{\pm}3.1$ & $36.0{\pm}2.9$ & $41.0{\pm}2.8$ & $4.6{\pm}0.8$ & $36.9{\pm}7.9$ & $17.4{\pm}1.6$ \\
NinaPro DB2      & $35.6{\pm}4.2$ & $32.3{\pm}2.9$ & $1.6{\pm}0.6$ & $3.7{\pm}0.8$ & $10.7{\pm}1.4$ & $46.3{\pm}2.1$ & $\mathbf{46.6{\pm}2.1}$ & $16.9{\pm}1.6$ & $21.5{\pm}6.3$ & $9.2{\pm}1.1$ \\
NinaPro DB3      & $22.4{\pm}2.7$ & $9.4{\pm}1.0$ & $0.8{\pm}0.1$ & $1.8{\pm}0.4$ & $6.4{\pm}0.9$ & $\mathbf{24.0{\pm}2.6}$ & $21.1{\pm}2.8$ & $0.2{\pm}0.1$ & $7.5{\pm}2.9$ & $2.1{\pm}0.5$ \\
NinaPro DB4      & $26.8{\pm}3.1$ & $24.6{\pm}2.6$ & $0.7{\pm}0.5$ & $3.3{\pm}1.3$ & $10.0{\pm}3.6$ & $32.0{\pm}8.5$ & $\mathbf{38.8{\pm}3.6}$ & $0.3{\pm}0.1$ & $8.0{\pm}1.1$ & $6.1{\pm}1.0$ \\
NinaPro DB5      & $13.8{\pm}2.1$ & $10.8{\pm}2.3$ & $0.9{\pm}0.9$ & $2.8{\pm}0.7$ & $4.8{\pm}1.3$ & $\mathbf{17.6{\pm}5.4}$ & $16.3{\pm}3.0$ & $0.9{\pm}0.2$ & $6.2{\pm}2.5$ & $3.7{\pm}0.3$ \\
NinaPro DB6      & $40.0{\pm}1.5$ & $31.1{\pm}2.1$ & $11.0{\pm}2.0$ & $14.2{\pm}1.2$ & $37.9{\pm}3.6$ & $41.0{\pm}5.1$ & $\mathbf{42.1{\pm}4.8}$ & $20.3{\pm}1.3$ & $18.5{\pm}4.2$ & $23.4{\pm}1.1$ \\
NinaPro DB7      & $57.8{\pm}2.3$ & $46.2{\pm}5.9$ & $2.2{\pm}0.9$ & $5.0{\pm}0.6$ & $15.9{\pm}2.5$ & $\mathbf{59.3{\pm}5.0}$ & $54.8{\pm}3.2$ & $2.1{\pm}0.7$ & $23.2{\pm}4.3$ & $9.6{\pm}1.0$ \\
GRABMyo          & $58.0{\pm}7.2$ & $65.7{\pm}8.8$ & $6.4{\pm}1.3$ & $12.8{\pm}1.6$ & $30.9{\pm}3.0$ & $65.7{\pm}1.6$ & $\mathbf{74.2{\pm}3.7}$ & $10.1{\pm}4.3$ & $22.9{\pm}2.8$ & $30.4{\pm}1.9$ \\
CapgMyo          & $\mathbf{27.2{\pm}2.6}$ & $17.9{\pm}8.4$ & $2.8{\pm}0.1$ & $10.5{\pm}3.5$ & $15.3{\pm}5.7$ & $17.0{\pm}5.4$ & $10.0{\pm}3.3$ & $11.3{\pm}3.7$ & $6.7{\pm}5.5$ & $3.0{\pm}0.8$ \\
Pinch            & $23.4{\pm}6.8$ & $16.0{\pm}7.3$ & $5.1{\pm}3.6$ & $13.9{\pm}3.8$ & $13.0{\pm}5.9$ & $\mathbf{38.1{\pm}9.2}$ & $26.8{\pm}5.5$ & $12.1{\pm}3.1$ & $21.0{\pm}6.4$ & $33.3{\pm}0.4$ \\
Roshambo         & $42.3{\pm}5.6$ & $\mathbf{57.6{\pm}8.3}$ & $38.1{\pm}11.4$ & $29.8{\pm}4.6$ & $21.9{\pm}8.4$ & $37.4{\pm}13.9$ & $55.0{\pm}13.0$ & $37.2{\pm}16.2$ & $54.5{\pm}1.9$ & $34.4{\pm}9.9$ \\
emg2pose         & $15.1{\pm}1.2$ & $12.7{\pm}2.1$ & $6.9{\pm}0.6$ & $15.8{\pm}1.0$ & $15.5{\pm}0.9$ & $\mathbf{71.2{\pm}0.6}$ & $67.5{\pm}0.5$ & $10.7{\pm}0.7$ & $12.7{\pm}0.6$ & $12.3{\pm}2.2$ \\
3DC              & $30.0{\pm}5.1$ & $48.8{\pm}8.4$ & $9.0{\pm}1.1$ & $19.3{\pm}3.6$ & $33.0{\pm}4.7$ & $\mathbf{62.1{\pm}3.6}$ & $59.2{\pm}4.4$ & $19.4{\pm}5.3$ & $16.4{\pm}3.3$ & $7.5{\pm}2.1$ \\
ContinuousTrans  & $65.8{\pm}3.9$ & $62.6{\pm}2.0$ & $27.9{\pm}2.0$ & $33.8{\pm}3.3$ & $73.3{\pm}3.6$ & $91.0{\pm}2.2$ & $\mathbf{92.9{\pm}1.1}$ & $45.7{\pm}9.3$ & $60.8{\pm}3.4$ & $56.8{\pm}0.7$ \\
Arm Translation  & $\mathbf{72.0{\pm}2.4}$ & $69.6{\pm}5.5$ & $28.1{\pm}4.1$ & $28.8{\pm}1.8$ & $67.0{\pm}8.6$ & $62.0{\pm}5.6$ & $64.8{\pm}5.2$ & $22.0{\pm}9.4$ & $41.0{\pm}2.6$ & $32.2{\pm}1.2$ \\
\midrule
HD-EMG           & $\mathbf{55.5{\pm}2.2}$ & $43.2{\pm}5.1$ & $33.0{\pm}3.9$ & $30.7{\pm}7.1$ & $38.1{\pm}9.6$ & $46.8{\pm}8.7$ & $47.8{\pm}6.6$ & $26.2{\pm}10.9$ & $34.6{\pm}4.0$ & $35.3{\pm}2.9$ \\
SIAT-LLMD        & $\mathbf{61.1{\pm}1.4}$ & $57.6{\pm}2.0$ & $9.3{\pm}1.6$ & $15.5{\pm}2.0$ & $22.7{\pm}0.7$ & $18.6{\pm}3.4$ & $27.1{\pm}1.6$ & $9.7{\pm}2.5$ & $25.9{\pm}1.1$ & $35.1{\pm}1.7$ \\
Mendeley Knee    & $82.4{\pm}1.7$ & $\mathbf{83.2{\pm}2.2}$ & $38.8{\pm}4.6$ & $44.7{\pm}4.5$ & $60.9{\pm}4.2$ & $52.8{\pm}3.8$ & $59.8{\pm}5.2$ & $48.8{\pm}5.8$ & $66.9{\pm}5.0$ & $53.3{\pm}0.4$ \\
MyPredict        & $38.4{\pm}2.0$ & $\mathbf{41.4{\pm}2.3}$ & $8.3{\pm}2.9$ & $16.3{\pm}1.8$ & $13.7{\pm}0.6$ & $16.9{\pm}1.2$ & $18.0{\pm}0.6$ & $18.4{\pm}1.6$ & $31.3{\pm}2.6$ & $13.3{\pm}0.6$ \\

\bottomrule
\end{tabular}
}
\end{table*}

\subsection{RQ4: Performance Consistency Across EMG Tasks}

\label{app:rq4}

RQ4 examines whether model performance remains consistent across upper-limb classification, lower-limb classification, and continuous EMG-to-text decoding. Table~\ref{tab:app_seq2seq} reports character error rate on emg2qwerty, for which lower values indicate better performance. Classification results are summarized from the FFT performance across the 15 upper-limb and four lower-limb datasets. The mean model ranks for upper- and lower-limb classification have Spearman correlations of $\rho=0.97$ for accuracy and $\rho=0.96$ for macro-F1. Performance on emg2qwerty is also associated with classification performance: its correlations with upper-limb classification are $\rho=0.85$ for both accuracy and macro-F1, while the corresponding correlations with lower-limb classification are $\rho=0.78$ and $\rho=0.83$, respectively. These results support the cross-task comparisons reported in Section~\ref{sec:rq4}.

\section{Detailed Dataset Descriptions}
\label{app:datasets}

This section provides comprehensive descriptions of all 20 datasets in~\methodName, including data collection protocols, hardware specifications, task paradigms, and subject demographics. We provide preprocessed versions of all datasets in a unified format on Hugging Face for reproducibility.

\begin{table*}[t]
\centering
\footnotesize
\caption{Summary of EMG datasets used in this benchmark.}
\label{tab:datasets}
\resizebox{0.85\textwidth}{!}{
\begin{tabular}{llccc}
\toprule
\textbf{Dataset} & \textbf{Sensor Location} & \textbf{Fs (Hz)} & \textbf{Channels} & \textbf{Task} \\
\midrule
\multicolumn{5}{l}{\textit{Hand/Arm Gesture Datasets}} \\
NinaPro DB1~\cite{atzori2014electromyography} & Forearm & 100 & 10 & Hand Gesture Classification \\
NinaPro DB2~\cite{atzori2014electromyography} & Forearm & 2000 & 12 & Hand Gesture Classification \\
NinaPro DB3~\cite{atzori2014electromyography} & Forearm & 2000 & 12 & Hand Gesture Classification \\
NinaPro DB4~\cite{pizzolato2017comparison} & Forearm & 2000 & 12 & Hand Gesture Classification \\
NinaPro DB5~\cite{pizzolato2017comparison} & Forearm & 200 & 16 & Hand Gesture Classification \\
NinaPro DB6~\cite{palermo2017repeatability} & Forearm & 2000 & 14 & Hand Gesture Classification \\
NinaPro DB7~\cite{krasoulis2017improved} & Forearm & 2000 & 12 & Hand Gesture Classification \\
GRABMyo~\cite{pradhan2022multi} & Forearm & 2048 & 28 & Hand Gesture Classification \\
CapgMyo~\cite{dai2021capg} & Forearm & 1000 & 128 & Hand Gesture Classification \\
Pinch~\cite{donati2018processing} & Forearm & 200 & 8 & Pinch Gesture Classification  \\
Roshambo~\cite{donati2018processing} & Forearm & 200 & 8 & Hand Gesture Classification \\
Arm Translation~\cite{kyranou2025emg} & Forearm & 2000 & 16 & Hand Gesture Classification \\
emg2pose~\cite{salter2024emg2pose} & Bilateral Wrists & 2000 & 16 & Behavioral-Stage Classification   \\
3DC~\cite{eddy2023libemg} & Forearm & 1000 & 10 & Hand Gesture Classification \\
ContinuousTrans~\cite{eddy2023libemg} & Hand & 2000 & 6 & Hand Gesture Classification \\
\midrule
\multicolumn{5}{l}{\textit{Leg/Locomotion Datasets}} \\
Mendeley Knee~\cite{mohr2019classification} & Leg & 2400 & 10 & Knee Injury Classification \\
HD-EMG~\cite{dimitrov2023high} & Leg & 2000 & 64 & Gait Classification \\
SIAT-LLMD~\cite{wei2023surface} & Leg & 1920 & 9 & Gait Phase Classification \\
MyPredict~\cite{schulte2023roessingh} & Leg & 1000 & 16 & Natural Locomotion \\
\midrule
\multicolumn{5}{l}{\textit{Sequence-to-Sequence Datasets}} \\
emg2qwerty~\cite{sivakumar2024emg2qwerty} & Bilateral Wrists & 2000 & 32 & Typing (Seq2Seq) \\
\bottomrule
\end{tabular}
}
\end{table*}

\subsection{Upper-Limb Gesture Datasets}
\label{app:datasets_upper}

\subsubsection{NinaPro Family (DB1-7)}

The NinaPro (Non-Invasive Adaptive Prosthetics) database~\cite{atzori2014electromyography,pizzolato2017comparison,krasoulis2017improved,palermo2017repeatability,krasoulis2019effect} constitutes the core of our upper-limb corpus, providing standardized hand and forearm gesture recordings across multiple databases:

\noindent\textbf{NinaPro DB1~\cite{atzori2014electromyography}:} Forearm placement, 100 Hz sampling, 10 channels, 27 subjects performing 52 gestures including finger movements, wrist rotations, and grasping patterns. Each gesture repeated 10 times with 5-second duration.

\noindent\textbf{NinaPro DB2~\cite{atzori2014electromyography}:} Forearm placement, 2000 Hz sampling, 12 channels, 40 able-bodied subjects performing 49 movements spanning finger and wrist motions, grasping and functional movements, and force-varying patterns.

\noindent\textbf{NinaPro DB3~\cite{atzori2014electromyography}:} Forearm placement, 2000 Hz sampling, 12 channels, 11 subjects with transradial amputations following the DB2 protocol. DB3 is one dataset in the benchmark recorded from an amputee population, and therefore the setting in which transfer is evaluated for users whose residual musculature differs structurally from the able-bodied majority.

\noindent\textbf{NinaPro DB4~\cite{pizzolato2017comparison}:} Forearm placement, 2000 Hz sampling, 12 channels, 10 able-bodied subjects performing 52 movements. DB4 and DB5 share an acquisition protocol and differ only in hardware, which makes the pair a controlled contrast between sampling rate and channel count.

\noindent\textbf{NinaPro DB5~\cite{pizzolato2017comparison}:} Forearm placement, 200 Hz sampling, 16 channels from two Myo armbands, 10 able-bodied subjects performing 52 movements under the same protocol as DB4. The low sampling rate and consumer-grade hardware test robustness to reduced temporal resolution.

\noindent\textbf{NinaPro DB6~\cite{palermo2017repeatability}:} Forearm placement, 2000 Hz sampling, 14 channels, 10 able-bodied subjects recorded across multiple sessions spanning five days, which supports intra-subject temporal robustness evaluation.

\noindent\textbf{NinaPro DB7~\cite{krasoulis2017improved}:} Forearm placement, 2000 Hz sampling, 12 channels, 20 able-bodied subjects and 2 subjects with amputations performing 41 movements.

For all NinaPro datasets, we retain the complete set of annotated gestures. This differs from prior evaluations that select smaller gesture subsets from NinaPro. The resulting higher-cardinality classification tasks are more challenging, and their absolute accuracies should therefore not be directly compared with results reported on reduced class subsets.

\subsubsection{Pinch \& Roshambo}
The Pinch and Roshambo datasets~\cite{donati2018processing} were recorded using a Myo armband with eight equally spaced sEMG sensors positioned around the middle of the forearm at 200 Hz. The Pinch dataset contains recordings from 22 subjects performing four pinch gestures between the thumb and the index, middle, ring, or little finger. The Roshambo dataset contains recordings from 10 subjects performing three gestures: rock, paper, and scissors. Each subject completed three sessions, with five repetitions of each gesture per session. 

\subsubsection{GRABMyo}

GRABMyo~\cite{pradhan2022multi} provides multi-session forearm and wrist sEMG recordings collected at 2048 Hz from 43 participants using 28 channels (16 forearm and 12 wrist channels). Participants performed 16 static hand and finger gestures and a rest condition, yielding 17 classes, with seven repetitions per class across three sessions conducted on separate days. This multi-day design supports evaluation of robustness to session-level and day-to-day variation. We process all 3 sessions for this dataset and use session 1 for evaluation.

\subsubsection{CapgMyo}

CapgMyo~\cite{dai2021capg} features ultra-high-density spatial sampling: forearm placement, 1000 Hz sampling, 128 channels in an 8×16 grid covering extrinsic hand and finger muscles. 18 subjects perform 8 isometric gestures with 3 recording sessions per subject. Because each CapgMyo trial is approximately 1 second long, applying the standard 3-second segmentation would discard the trial. We therefore use each complete trial as one sample and temporally resample it to 3000 points, preserving the trial boundary while matching the input dimensionality used by the other datasets.

\subsubsection{3DC}

3DC~\cite{eddy2023libemg} features forearm placement, 1000 Hz sampling, 10 channels, 22 subjects performing 11 hand gesture classes. Moderate channel count and subject pool provide a representative mid-scale benchmark for cross-subject generalization.

\subsubsection{ContinuousTrans}

ContinuousTrans~\cite{eddy2023libemg} features forearm placement, 2000 Hz sampling, 6 channels, 43 subjects performing 7 gesture classes including continuous transition states between gestures. The low channel count and transition-inclusive labeling test model robustness under sparse spatial sensing and temporally ambiguous class boundaries.

\subsubsection{Arm Translation}

Arm Translation~\cite{kyranou2025emg} targets upper-limb gesture recognition with forearm placement (two rows of eight Delsys Trigno electrodes), 2000 Hz sampling, 16 channels, and 8 subjects performing 6 grasp classes (power, lateral, pointer, tripod, open, and rest). Despite the small subject count, the dataset is distinctive in that each grasp is recorded at nine arm positions on a $3\times3$ spatial grid across two days, so it captures limb-position-induced variability in the EMG signal under a fixed electrode placement rather than a different recording site.

\subsubsection{emg2pose}

emg2pose~\cite{salter2024emg2pose} is a large-scale dataset targeting hand pose estimation. Bilateral wrist placement, 2000 Hz sampling, 16 channels, and 193 subjects (174 identity-disjoint training subjects and 19 held-out subjects) performing 29 behavioral stage categories. The large subject pool enables robust out-of-distribution generalization evaluation. In this benchmark, we repurpose the dataset as a 29-class behavioral-stage classification task using the original metadata stage field; the continuous joint-angle targets are not used.

\subsection{Lower-Limb Locomotion Datasets}
\label{app:datasets_lower}



\subsubsection{Mendeley Knee}
\label{sec:mendeley}

The Mendeley Knee dataset from Mohr et al.~\cite{mohr2019classification} contains bilateral lower-limb EMG recordings collected at 2400 Hz from five muscles per leg during treadmill walking at 4.5 km/h. Although 71 participants were initially recorded, 61 were retained in the released analysis set. We formulate a binary classification task distinguishing participants with a previous knee injury from those without a knee injury history. Because injury history is a participant-level label, protocols in which the same participants appear during training and testing may allow models to exploit participant-specific signal characteristics. The LP, FFT, scratch, and five-shot results should therefore be interpreted as performance on seen participants, whereas the held-out and LOSO protocols provide more informative estimates of injury-related generalization to unseen participants. We report all protocols for completeness. As a sensitivity check, excluding Mendeley Knee changes the upper- versus lower-limb rank correlation in Section~\ref{sec:rq4} from $0.97$ to $0.92$ for accuracy and from $0.96$ to $0.98$ for macro-F1, without changing the qualitative conclusion.

\subsubsection{HD-EMG}

Dimitrov et al.~\cite{dimitrov2023high} dataset provides high-density spatial sampling during locomotion: leg placement, 2000 Hz sampling, 64 channels covering tibialis anterior, peroneus, and gastrocnemius muscles. 10 subjects perform comprehensive activities including walking, running, stair climbing, and ramp negotiation across multiple speeds. High channel density enables spatial transfer evaluation. The released data are already segmented into individual gait cycles. Each gait-cycle segment is treated as one sample and reflect-padded to 3000 points when shorter than the target length.

\subsubsection{SIAT-LLMD}

Wei et al.~\cite{wei2023surface} dataset captures natural lower-limb motion intent: leg placement, 1920 Hz sampling, 9 channels. Multiple subjects and sessions enable cross-subject and cross-session evaluation for gait phase classification.

\subsubsection{MyPredict}

Schulte et al.~\cite{schulte2023roessingh} released the MyPredict database for lower-limb activity prediction, consisting of three subsets with different sEMG configurations. We process all the subsets and use MyPredict1 subset for evaluation, 16 bipolar channels (eight muscles per leg, bilateral), recorded with Delsys Trigno at 1000~Hz. Each subject performed 40 trials of a fixed circuit in which they transitioned freely between sitting, standing, level walking, stair ascent/descent, ramp ascent/descent, uneven terrain, and confined-space stepping, providing naturally segmented locomotion activities for classification. The multi-array subsets (MyPredict~2 and~3) use a different electrode layout and are not included.

\subsection{Sequence-to-Sequence Dataset}
\label{app:datasets_seq2seq}

\subsubsection{emg2qwerty}

Sivakumar et al.~\cite{sivakumar2024emg2qwerty} enable continuous typing decoding: bilateral wrist placement (32 channels total), 2000 Hz sampling. 108 subjects perform natural QWERTY keyboard typing. The dataset contains 5,262,671 keystrokes across 1,135 sessions (346.4 hours). Unlike discrete classification where each window contains one gesture, typing segments capture continuous muscle activation patterns corresponding to variable-length character sequences.

The dataset provides timestamped keypress events aligned with EMG recordings, enabling evaluation using CTC~\cite{graves2006connectionist}. We segment each recording into consecutive non-overlapping 3-second windows with a stride of 3 seconds. Models predict character sequences at each timestep, and CTC automatically learns temporal alignment. Evaluation uses Character Error Rate (CER) computed by concatenating predictions across windows to reconstruct full typed text.

\subsection{Unified Preprocessing Pipeline}
\label{app:preprocessing}

All datasets undergo a unified sampling-rate-aware preprocessing pipeline to ensure fair cross-dataset comparison despite severe hardware heterogeneity (100-2400 Hz sampling rates, 6-128 channels).

\noindent\textbf{Stages 1--2: Resampling and Bandpass Filtering.} For recordings acquired at or above 1000 Hz, raw signals are first resampled to 1000 Hz using polyphase filtering and then filtered using a 4th-order Butterworth bandpass filter with 20--450 Hz cutoffs. For recordings acquired below 1000 Hz, bandpass filtering is instead performed at the original sampling rate before upsampling, with the upper cutoff constrained to remain below the original Nyquist frequency. The filtered signals are then resampled to 1000 Hz. This sampling-rate-aware ordering standardizes temporal resolution without assuming or recovering frequency components unavailable in the original recordings.

\noindent\textbf{Stage 3: Notch Filtering.} Power-line interference is eliminated using notch filters at 50 Hz/60 Hz with quality factor $Q=30$ when the corresponding frequency is supported by the original sampling rate.

\noindent\textbf{Stage 4: Instance Normalization.} Raw EMG amplitudes vary substantially across subjects, electrode placements, acquisition devices, and recording conditions~\cite{merletti2016surface,besomi2019consensus}. To reduce these dataset- and subject-specific scale differences, each signal segment is independently Z-score normalized per channel. Window-level Z-score normalization has also been shown to support calibration-free EMG motion recognition and cross-subject generalization~\cite{tanaka2022sliding}. This normalization provides a consistent scale-invariant input interface across datasets without requiring target-user calibration statistics.

\noindent\textbf{Task-Specific Segmentation.}
We use non-overlapping 3-second windows for all downstream tasks (with the exception of CapgMyo and HD-EMG), providing a common physical context and a fixed input length of 3000 samples after resampling to 1\,kHz. This duration captures sustained gesture dynamics, multiple locomotion cycles, and short typing sequences while remaining computationally feasible for all evaluated backbones. Each model processes the resulting windows using its native tokenization. Classification windows carry the corresponding discrete trial or activity label. For emg2qwerty, timestamped keypresses within each window form a variable-length target sequence, and the models are trained using CTC loss~\cite{graves2006connectionist} without explicit frame-level alignment.

\noindent\textbf{Dataset Partitioning.}
We first divide the subjects in each classification dataset into two disjoint cohorts: Pool~A, containing approximately 90\% of the subjects, and Pool~B, containing the remaining 10\%. Within each pool, we construct training, validation, and test partitions separately for each subject using an approximately 70/15/15 split. When trial or repetition identifiers are available, complete trials or repetitions are assigned to the three partitions before windowing, and all windows from the same trial or repetition remain in the same partition. For datasets without reliable trial or repetition identifiers, each subject's recordings are ordered temporally and divided into contiguous 70\%, 15\%, and 15\% blocks before windowing. Thus, windows from the same trial or temporal block are not distributed across partitions. The Pool~A partitions are used for the in-distribution LP, FFT, and scratch protocols. For zero-calibration held-out-user evaluation, models trained on Pool~A are evaluated on the fixed Pool~B test partition without using Pool~B training data. For five-shot adaptation, labeled examples are sampled from the Pool~B training partition, while evaluation remains on the fixed Pool~B test partition. LOSO is constructed separately by holding out an entire subject for testing in each fold.

\section{Foundation Model Architectures \& Training}
\label{app:models}

This section provides detailed architectural specifications for each foundation model evaluated in our benchmark. All information has been verified against the original publications. Table~\ref{tab:model_summary} summarizes key characteristics of all evaluated models.

\begin{table*}[t]
\centering
\caption{
Summary of the seven model families and nine pretrained configurations
evaluated in \methodName. Parameter counts correspond to the checkpoints
used in our experiments.
}
\label{tab:model_summary}

\setlength{\tabcolsep}{3pt}
\renewcommand{\arraystretch}{1.10}
\small

\begin{tabular}{
@{}
>{\raggedright\arraybackslash}p{0.14\textwidth}
>{\raggedright\arraybackslash}p{0.23\textwidth}
>{\raggedright\arraybackslash}p{0.18\textwidth}
>{\raggedright\arraybackslash}p{0.28\textwidth}
>{\raggedleft\arraybackslash}p{0.08\textwidth}
@{}
}
\toprule
\textbf{Model} &
\textbf{Architecture} &
\textbf{Pretraining Objective} &
\textbf{Native Scope / Input} &
\textbf{Params.} \\
\midrule

MOMENT-S/B &
Encoder; patch tokens &
Masked-patch reconstruction &
Heterogeneous real-world numerical time series &
40M / 125M \\
\addlinespace[2.5pt]

Chronos-2-S/B &
Encoder; temporal and group attention &
Direct quantile forecasting &
Univariate and multivariate targets with optional covariates &
28M / 120M \\
\addlinespace[2.5pt]

TimesFM &
Decoder; patched input &
Multi-horizon forecasting &
Primarily univariate numerical time series &
200M \\
\addlinespace[2.5pt]

Lag-Llama &
Decoder; lagged covariates &
Autoregressive probabilistic forecasting &
Univariate series with lag and temporal covariates &
2.45M \\
\addlinespace[2.5pt]

PatchTST &
Encoder; channel-independent patching &
Forecasting &
General time series &
$\sim$1M \\
\addlinespace[2.5pt]

BRANT &
Temporal followed by spatial encoders &
Masked-signal reconstruction &
Multichannel iEEG/SEEG with time--frequency features &
$>$500M \\
\addlinespace[2.5pt]

UniTS &
Unified Transformer with task tokenization &
Multi-task modeling &
Forecasting, classification, imputation, and anomaly detection &
3.4M \\

\bottomrule
\end{tabular}
\end{table*}




\subsection{MOMENT}

MOMENT (MOdel for Masked tiME-series Analysis)~\cite{goswami2024moment} is a family of encoder-only transformer models pretrained via masked reconstruction on a large corpus of public time series datasets.

\textbf{MOMENT-Small.} This variant contains approximately 40M parameters and uses a standard transformer encoder architecture. The model is pretrained using a masked autoencoding objective where random patches of the input time series are masked and the model learns to reconstruct them.

\textbf{MOMENT-Base.} The larger variant contains approximately 125M parameters, using the same architectural design as MOMENT-Small but with increased model capacity.

\subsection{Chronos-2}

We evaluate two variants of Chronos-2~\cite{ansari2025chronos}, a family of encoder-only transformer models for time series forecasting.

\textbf{Chronos-2-S.} This model contains 28M parameters and uses an encoder-only transformer architecture inspired by the T5 encoder. The model introduces a \textit{group attention mechanism} that partitions the input sequence into groups and applies attention within each group, improving computational efficiency while maintaining forecasting accuracy. Chronos-2-S is pretrained on a diverse mixture of synthetic and real-world time series data using a forecasting objective.

\textbf{Chronos-2-B.} The larger variant contains 120M parameters, maintaining the same encoder-only architecture with group attention as Chronos-2-S. The increased capacity enables better performance on complex forecasting tasks while retaining the computational benefits of the group attention mechanism.


\subsection{TimesFM}

TimesFM~\cite{das2023decoder} is a decoder-only transformer model for univariate time series forecasting. The model contains 200M parameters and uses a standard decoder-only architecture similar to language models. TimesFM is pretrained on approximately 100 billion real-world timepoints from the Google Trends dataset and other sources, making it one of the largest-scale pretraining efforts for time series FMs.

\subsection{Lag-Llama}

Lag-Llama~\cite{rasul2310lag} is a decoder-only transformer model for probabilistic time series forecasting, inspired by the LLaMA language model architecture. The model contains 2,449,299 parameters (approximately 2.45M) and uses lags as input features, with a tokenization scheme that constructs lagged features from prior values of the time series. The model employs rotary positional embeddings (RoPE) and RMSNorm, following the LLaMA design. Lag-Llama is pretrained on 27 diverse time series datasets totaling approximately 352 million data windows.

\subsection{PatchTST}

PatchTST~\cite{nie2022time} is a patch-based transformer model that segments time series into subseries-level patches which serve as input tokens. The model uses a channel-independent design where each channel (univariate series) is processed independently through a shared transformer backbone. The number of parameters is approximately 1M, though this varies depending on configuration (number of layers, hidden dimensions, and number of attention heads). The patch-based tokenization significantly reduces computational complexity compared to point-wise transformers. 

\subsection{BRANT}

BRANT (Brain Transformer)~\cite{zhang2023brant} is a foundation model specifically designed for intracranial neural signals (iEEG). The model contains more than 500M parameters and employs a dual transformer encoder architecture: a temporal encoder that captures dependencies across time, and a spatial encoder that models correlations across different recording channels. BRANT processes signals in both time and frequency domains and is pretrained via masked patch prediction on approximately 1TB of iEEG data. An updated version, BRANT-2, extends the model to handle broader types of brain signals including scalp EEG.

\subsection{UniTS}

UniTS~\cite{gao2024units} is a unified multi-task time series model that handles forecasting, classification, imputation, and anomaly detection within a single framework. The model uses a modified transformer architecture with both sequence attention (across time) and variable attention (across channels), along with a dynamic linear operator for modeling complex temporal relationships. UniTS employs task tokenization to encode different task specifications into a unified representation. The model is pretrained on 38 datasets from diverse domains including human activity sensors, healthcare, engineering, and finance. 

\subsection{Training Hyperparameters}
\label{app:training}

We use a single downstream adaptation recipe across all backbones so that pretrained and randomly initialized models are evaluated under the same training conditions. The objective is to characterize relative transfer behavior under a standardized protocol, rather than to estimate the individually tuned performance ceiling of each model, consistent with fixed-protocol evaluation in prior EMG and EEG benchmarks~\cite{yang2024emgbench,kokate2026channel}.

For the comparison between pretrained full fine-tuning and training from scratch, each paired configuration used the same backbone architecture, input adapter, pooling operation, classification head, data splits, optimizer settings, maximum training budget, and validation-based checkpoint-selection procedure. The two settings differed only in backbone initialization: the pretrained setting initialized the backbone from the released checkpoint, whereas the scratch setting randomly initialized the same backbone. The input adapter and classification head were randomly initialized in both settings. This setup therefore compares pretrained and randomly initialized versions of the same architecture under an otherwise identical downstream training procedure.

All models were optimized using AdamW with $\beta_1=0.9$, $\beta_2=0.95$, weight decay $0.01$, and $\epsilon=10^{-7}$. Linear probing used a learning rate of $1\times10^{-5}$, while both pretrained full fine-tuning and scratch training used a learning rate of $6\times10^{-5}$. Models were trained for up to 50 epochs with validation-based early stopping, and the checkpoint achieving the best validation performance was used for test evaluation. Most scratch runs reached stable validation performance and selected their best checkpoint before the 50-epoch limit, suggesting that the maximum training horizon was not binding for the majority of configurations.

The batch size was 32 whenever memory permitted and was reduced to 4--16 for larger architectures, including MOMENT-Base and BRANT, to accommodate GPU memory constraints on 80\,GB A100/H100 and 24\,GB L4 GPUs. These batch-size differences were determined by memory requirements rather than model-specific tuning. Within each paired pretrained-versus-scratch comparison, the batch size was held fixed.

Five-shot adaptation used a fixed training budget of 30 epochs without early stopping or target-cohort validation-based checkpoint selection. The final-epoch checkpoint was evaluated on the fixed Pool~B test set. Five-shot adaptation was repeated with five random seeds (42--46). Each seed controls both the sampling of the five labeled examples per class from the Pool~B training split and the randomness of the adaptation run, while the Pool~A--Pool~B subject partition, the Pool~B test split, and the Pool~A checkpoint used as the starting point are held fixed across seeds. Reported five-shot results are the mean and standard deviation of macro-F1 across the five seeds.

\subsection{Downstream Interfaces}
\label{app:classification_interfaces}

Each preprocessed EMG window is represented as
$\mathbf{x}\in\mathbb{R}^{C\times3000}$, where $C$ denotes the dataset-specific number of channels.
Because the evaluated foundation models expose different native input and representation interfaces,
each window is arranged according to the input format expected by the corresponding backbone.
The resulting latent representations are aggregated into a fixed-dimensional sample-level representation
and mapped to $K$ class logits using a single linear classification layer, where $K$ denotes the number
of classes in the corresponding dataset. All classification models are optimized using cross-entropy loss.
Except for MOMENT, which uses channel-wise concatenation following temporal pooling, all evaluated
backbones use mean aggregation across channels. For each model, the input interface, representation
aggregation, and classification-head architecture are held fixed across evaluation protocols.

Under linear probing, all backbone parameters are frozen and only the final linear classification layer is trained.
Under full fine-tuning, the complete model, including its input-embedding modules and backbone, is updated.
For each architecture, the input interface, pooling operation, and classification-head architecture are held fixed
across pretrained and randomly initialized variants. The pretrained variant initializes the available backbone
components from the released checkpoint, whereas the scratch variant initializes the same components randomly.

\noindent\textbf{MOMENT.}
The EMG window is passed directly to MOMENT's native multichannel input interface, where each channel
is patchified and encoded using the shared backbone. Patch representations are averaged over the temporal
dimension, concatenated across channels, and mapped to the class logits using a linear classification layer.

\noindent\textbf{Chronos-2.}
Chronos-2 natively supports grouped multivariate time series. We treat the EMG channels within each window
as related variates and assign them a shared group identifier, while channels from different windows are assigned
to different groups. Each channel is processed using Chronos-2's native normalization, patching, and
input-embedding modules, followed by its temporal and group-attention encoder. The resulting context-patch
representations are averaged first across the patch dimension and then across channels, producing a fixed-dimensional
representation that is mapped to the class logits using a linear classification layer.

\noindent\textbf{TimesFM.}
Because TimesFM is designed for univariate time series, each EMG channel is treated as an independent series
and processed using the model's native patching and input-embedding modules followed by the shared pretrained
transformer backbone. Representations are averaged over the temporal-patch dimension and then averaged across
channels before being mapped to the class logits using a linear classification layer.

\noindent\textbf{Lag-Llama.}
Because Lag-Llama is designed for univariate time series, each EMG channel is treated as an independent series
and processed using its native lag-feature construction and shared transformer backbone. Representations are
averaged over the temporal dimension and then across channels before being mapped to the class logits using a
linear classification layer.

\noindent\textbf{PatchTST.}
The input is transposed to the time-major layout expected by PatchTST. Each channel is patchified and encoded
as an independent variable stream using the shared transformer backbone. The resulting representations are
averaged across the patch and channel dimensions and mapped to the class logits using a linear classification layer.

\noindent\textbf{BRANT.}
The EMG window is processed using BRANT's native temporal and channel encoders, which model temporal
dependencies and cross-channel interactions, respectively. The resulting representations are averaged across
the temporal-patch and channel dimensions and mapped to the class logits using a linear classification layer.

\noindent\textbf{UniTS.}
The EMG channels are provided through UniTS's native variate dimension and processed using its native prompt
and classification tokens. We extract the pooled classification representation before the model's category-token
matching operation, average it across channels, and map the resulting representation to the class logits using a
linear classification layer.

\textbf{Sequence-decoding interface for emg2qwerty.}
Each window is processed by the corresponding backbone to produce a sequence of temporal representations, whose length depends on the backbone architecture. At each temporal position, representations from the 32 EMG channels are concatenated, yielding a $32d_{\text{model}}$-dimensional feature. A randomly initialized linear projection
$\mathbb{R}^{32d_{\text{model}}}\to\mathbb{R}^{71}$
is then applied independently at each temporal position to produce logits over 70 character classes and the CTC blank symbol (index 0). The decoding head has the same design across backbones. Models are trained using \texttt{torch.nn.CTCLoss} with blank index 0, \texttt{zero\_infinity=True}, and mean reduction, applied to the log-softmax of the temporal logits. The CTC input length is determined by the temporal resolution of the corresponding backbone output; no class weighting or auxiliary regularization is used. Decoding is greedy: the argmax class is selected at each temporal position, consecutive repeated symbols are collapsed, and blank symbols are removed.

\subsection{Conventional Supervised Reference}
\label{app:lstm}

Following the convolutional--recurrent design of Lee et al.~\cite{lee2023lstm}, we include a bidirectional LSTM (BiLSTM) as a representative generic supervised temporal reference. We adapt the architecture to the unified multichannel input format used in our benchmark. The model first applies a channel-wise depthwise 1D convolution to reduce each input sequence to approximately 150 time steps, followed by a linear projection that jointly maps the input channels to a 128-dimensional representation at each time step. Two stacked bidirectional LSTM layers, each with a hidden size of 256 per direction and a dropout rate of 0.4, produce a 512-dimensional sequence representation, which is passed to a linear classification head. The resulting model contains approximately 2.4M parameters, with minor variation across datasets due to differences in the numbers of input channels and output classes. The BiLSTM is randomly initialized and trained using the same data splits, validation-based checkpoint selection, and supervised evaluation protocols as the other train-from-scratch models. We report its performance under the scratch, held-out-user, LOSO, and five-shot protocols. It is not included in the LP-versus-FFT comparison because these settings require a pretrained checkpoint: LP freezes the pretrained backbone, whereas FFT updates it during downstream adaptation. The BiLSTM serves as a representative conventional supervised reference rather than an exhaustive comparison with task-specific EMG architectures.

\end{document}